\documentclass[10pt,twocolumn,letterpaper]{article}
\usepackage[pagenumbers]{cvpr}

\usepackage{iftex}
\ifXeTeX
  \usepackage[no-math]{fontspec}
  \IfFontExistsTF{Times New Roman}
    {\setmainfont{Times New Roman}}
    {\setmainfont{TeX Gyre Termes}}
\fi
\usepackage{amsthm}
\usepackage{microtype}
\usepackage{xr-hyper}

\AtBeginDocument{\crefname{proposition}{Proposition}{Propositions}}

\makeatletter
\providecommand{\@LN@col}[1]{}
\providecommand{\@LN}[2]{}
\makeatother

\definecolor{cvprblue}{rgb}{0.21,0.49,0.74}
\usepackage[pagebackref,breaklinks,colorlinks,allcolors=cvprblue]{hyperref}
\def\confName{CVPR}
\def\confYear{2027}
\title{RefCloth: Transferable Adversarial Tablecloths for Vision-Language-Action Models}

\author{
Shaozhen Liu\\
School of Artificial Intelligence, University of Chinese Academy of Sciences\\
Institute of Automation, Chinese Academy of Sciences\\
{\tt\small liushaozhen2025@ia.ac.cn}
\and
Zhiwei Li \quad
Jiayang Sun \quad
Shuangjun Du \quad
Yi Zhong  \quad
Shenrui Ma  \quad
Yaozu Liu  \quad
Qi Li\thanks{Correspondence to \texttt{qli@nlpr.ia.ac.cn} and \texttt{znsun@nlpr.ia.ac.cn}} \quad
Zhenan Sun\footnotemark[1]\\
University of Chinese Academy of Sciences\\
}

\begin{document}
\makeatletter
\apptocmd{\@maketitle}{\begin{minipage}{\textwidth}
\centering
\includegraphics[width=.98\textwidth]{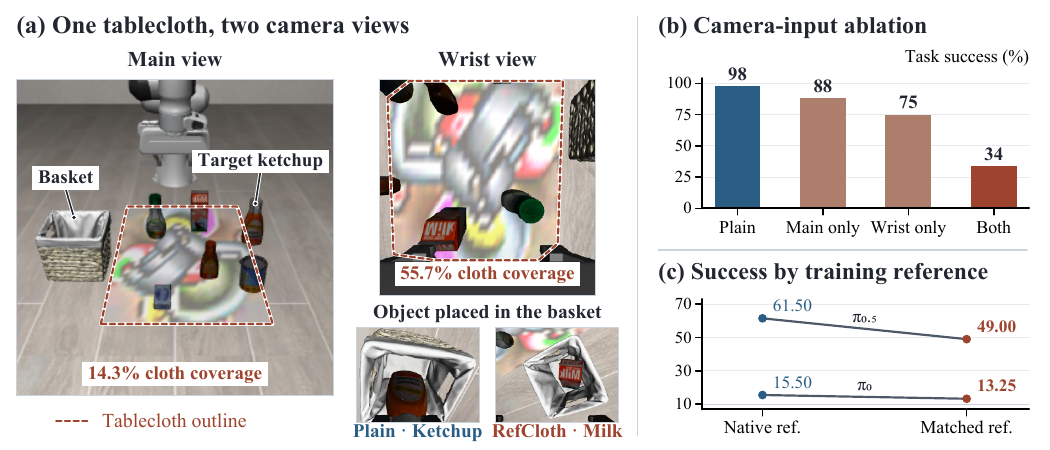}
\captionsetup{hypcap=false}
\captionof{figure}{\textbf{RefCloth at a glance.}
(a) Simultaneous main and wrist views of the same LIBERO-Object state.
The outlines and coverage percentages are derived from same-state color-marker renders.
In paired rollouts, Plain places ketchup in the basket and RefCloth places milk.
(b) Closed-loop task success for \(\pi_{0.5}\) on 100 paired Object episodes per camera condition.
The labels indicate which views receive RefCloth, with Plain in the remaining views.
(c) Task success in the native simulator across four LIBERO suites, with 400 episodes per policy and training reference.}
\label{fig:motivation}
\end{minipage}
\par\vspace{4pt}
}{}{\errmessage{Unable to place teaser after title}}
\makeatother
\maketitle
\begin{abstract}
Closed-loop vision-language-action (VLA) policies rely continuously on multi-view cameras, yet existing visual attacks typically hijack sensor streams with 2D patches or alter target object meshes that require retraining when manipulation objects change. We introduce RefCloth, a framework that optimizes one persistent adversarial tablecloth on a frozen source policy and deploys it zero-shot across unseen tasks and target models without modifying task objects or cameras. Throughout manipulation, static third-person and moving wrist views observe this shared world-bound texture under evolving occlusions. To generalize across shifting surface views while bridging the gap between differentiable and native renderers, RefCloth combines bounded texture parameterization with shared surface UV transformations and a same-path clean reference that removes static rendering offsets at initialization. Across four LIBERO suites in native simulation, one frozen texture transfers from OpenVLA-OFT to $\pi_0$ and $\pi_{0.5}$, reducing success relative to an unoptimized cloth by 62.50 and 43.75 percentage points on average and far exceeding amplitude- and variation-matched random controls. Dual-view audits and target closed-loop camera interventions show that moving wrist exposure drives over twice the standalone drop of the main view with strong synergy, while ablations confirm the roles of UV diversification and path calibration.
\end{abstract}
\section{Introduction}
\label{sec:intro}
Vision-language-action (VLA) models map visual observations and language instructions to robot control~\cite{rt2,openvla}, with large-scale data~\cite{oxe,droid} and pretrained representations~\cite{palme,siglip,dinov2} enabling generalist policies~\cite{octo,spatialvla} across transformer~\cite{rt1,rt2,act}, open VLA~\cite{openvla,oft}, and diffusion or flow-based architectures~\cite{diffusionpolicy,rdt1b,pi0,pi05}. Across these diverse action designs, closed-loop execution relies continuously on multi-view camera streams as the arm moves. This reliance exposes VLA policies to visual attacks via image patches~\cite{roboticattack,upa,edpa,tapdreamer}, object textures~\cite{tex3d,unitexture}, and scene perturbations~\cite{trap}. Yet the practical threat of an attack hinges on two questions, how much modification privilege the attacker needs, and whether one perturbation transfers across tasks, objects, and models without retraining.

Existing attack carriers struggle to satisfy both criteria. Camera-coordinate patches~\cite{patch,upa,edpa} require privileged access to hijack live sensor streams, whereas object-bound textures~\cite{meshadv,tex3d,unitexture} and CoT surfaces~\cite{trap} must alter target meshes or retrain for new objects, tasks, or policies while occupying minimal pixel area (\cref{fig:deployment}a). Instead, we investigate a passive workspace tablecloth that requires neither sensor intrusion nor object modification, remaining fixed and reusable without retraining across tasks and policies. Because third-person and moving wrist cameras continuously observe the workspace under evolving occlusions, one world-bound cloth projects into both views with substantial pixel coverage, misleading closed-loop object selection (\cref{fig:motivation}a). Crucially, as the gripper approaches targets, large wrist-camera exposure drives over twice the standalone disruption of the third-person view and acts synergistically when both views observe the cloth (\cref{fig:motivation}b).

Learning one frozen workspace texture on a source policy for zero-shot native deployment poses two technical challenges. First, because the cloth is spatially decoupled from target objects, varied reaching trajectories expose shifting, partially occluded local sub-regions across the two cameras. Second, its large footprint amplifies static synthesis discrepancies between differentiable~\cite{softras,nvdiffrast,pytorch3d} and native renderers, inducing 8.58--25.74\% source-feature offsets even on an unoptimized Plain cloth (\cref{fig:deployment}b). We introduce \textbf{RefCloth} to address both challenges on a frozen source policy. Shared rotations, reflections, and translations in surface UV coordinates~\cite{eot} force the bounded control grid to remain disruptive across arbitrary local surface slices and dual-view projections rather than overfitting to training trajectories. Simultaneously, routing the unperturbed base cloth through the identical synthesis path eliminates the static zero-point renderer offset at initialization, substantially improving zero-shot transfer across target policies over an uncalibrated native reference (\cref{fig:motivation}c). One exported PNG is then deployed frozen across unseen tasks and target models.

Our evaluation separates carrier placement from texture optimization across four LIBERO suites~\cite{libero} as one frozen texture transfers from OpenVLA-OFT~\cite{oft} to $\pi_0$~\cite{pi0} and $\pi_{0.5}$~\cite{pi05}. Our main contributions are threefold.
\begin{itemize}
\item \textbf{Low-privilege, zero-retraining 3D scene carrier.} To our knowledge, we are the first to optimize a single frozen, object-agnostic 3D tablecloth texture that transfers zero-shot across tasks and target VLA policies without altering objects, hijacking sensors, or retraining.
\item \textbf{Surface-augmented, path-calibrated optimization.} We combine shared dual-view UV transformations with a same-path clean feature baseline that eliminates static cross-renderer offsets during source optimization.
\item \textbf{Theoretical and empirical dual-view analysis.} We provide the first formal characterization of cross-renderer baseline drift and local dual-view sensitivity for workspace carriers, validated via de-confounded random controls and closed-loop camera interventions.
\end{itemize}

\begin{figure*}[t]
\centering
\includegraphics[width=\textwidth]{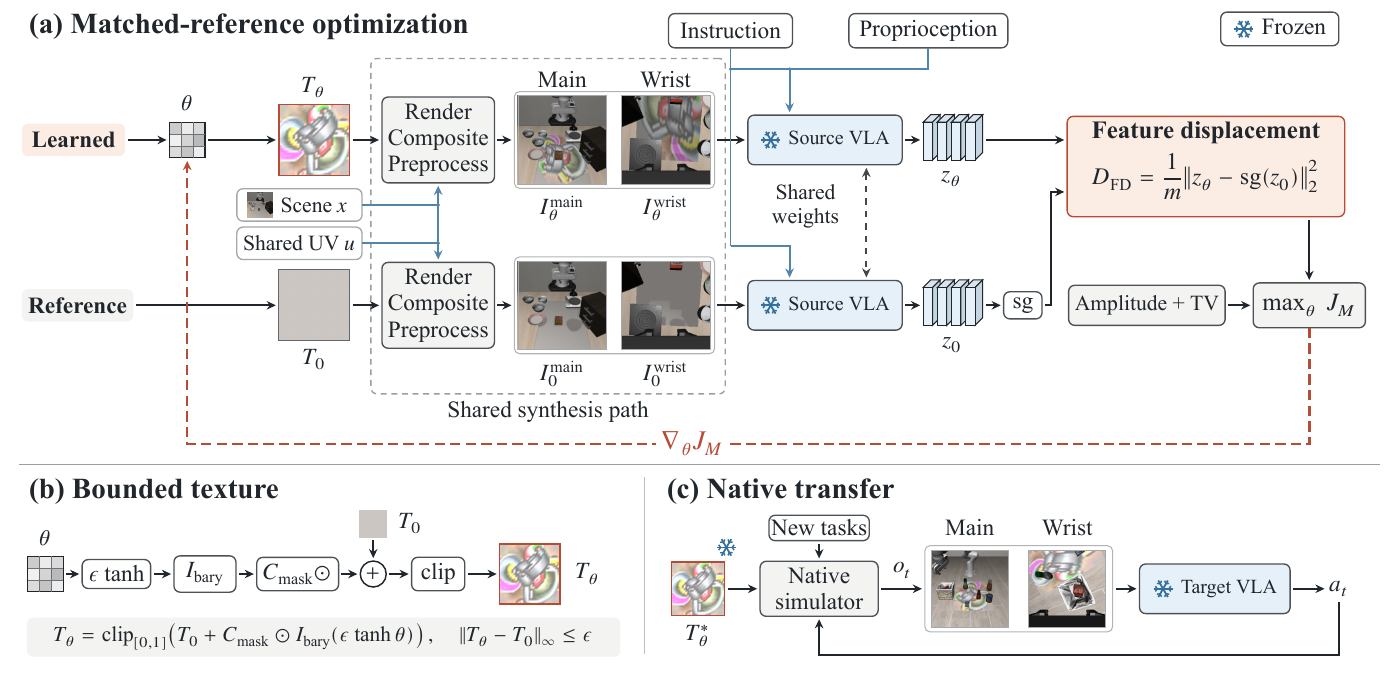}
\caption{\textbf{RefCloth optimization and native closed-loop transfer.}
(a) Both branches share scene state $x$, UV transformation $u$, the differentiable rasterization path, and frozen source VLA weights. Detached path-consistent feature displacement and amplitude/TV penalties optimize texture grid $\theta$. Images are $224\times224$ source-path renderings.
(b) Bounded texture generation via barycentric interpolation and clipping.
(c) Fixed-PNG zero-shot deployment in native simulation across unseen tasks and target policies, shown with Object camera views.}
\label{fig:method}
\end{figure*}
\begin{table}[t]
\centering\footnotesize
\caption{\textbf{Comparison with representative VLA visual attacks.} Columns indicate object-agnostic carrier (Obj.-agn.), 3D-consistent dual-view projection (3D-Dual), single frozen texture reuse (Reuse), optimization without prescribed action targets or CoT traces (No-Tgt.), and de-confounded carrier/random controls (Ctrl.).}
\label{tab:related_comparison}
\resizebox{\columnwidth}{!}{%
\setlength{\tabcolsep}{2.2pt}
\begin{tabular}{@{}llccccc@{}}
\toprule
Method & Carrier & Obj.-agn. & 3D-Dual & Reuse & No-Tgt. & Ctrl.\\
\midrule
UPA~\cite{upa}, EDPA~\cite{edpa} & 2D patch & \checkmark & -- & \checkmark & \checkmark & --\\
Tex3D~\cite{tex3d} & 3D mesh & -- & -- & -- & \checkmark & --\\
UniTexture~\cite{unitexture} & 3D mesh & -- & \checkmark & \checkmark & -- & --\\
TRAP~\cite{trap} & CoT cloth & \checkmark & -- & -- & -- & --\\
\textbf{RefCloth (Ours)} & \textbf{3D cloth} & \checkmark & \checkmark & \checkmark & \checkmark & \checkmark\\
\bottomrule
\end{tabular}%
}
\end{table}

\section{Related Work}
\label{sec:related}
\paragraph{Generalist visuomotor policies and robustness.}
Language-conditioned robotics spans skill selection~\cite{saycan}, visual and multimodal representations~\cite{palme,siglip,dinov2}, large-scale datasets~\cite{oxe,droid}, and generalist policies~\cite{rt1,rt2,octo}. OpenVLA~\cite{openvla}, its fine-tuning recipe~\cite{oft}, and SpatialVLA~\cite{spatialvla} provide accessible VLA backbones, whereas diffusion~\cite{diffusionpolicy,rdt1b}, action-chunking~\cite{act}, and flow-matching policies~\cite{pi0,pi05,xvla} adopt continuous action heads, motivating visual attacks agnostic to target action spaces. Because closed-loop execution also depends on scene conditions, evaluation alongside LIBERO~\cite{libero} must account for simulation fidelity and environmental variation~\cite{simpler,liberoplus,robotwin,colosseum,maniskill3}, motivating controls that isolate adversarial optimization.

\paragraph{Universal and physical visual attacks.}
Gradient-based attacks~\cite{fgsm,pgd,cw} extend to universal perturbations~\cite{uap} and patches~\cite{patch,lavann}, while physical attacks model viewpoint shifts~\cite{rp2,advtshirt} and 3D appearance~\cite{meshadv,fca,das}. To improve transfer across views and models, expectation over transformations~\cite{eot} optimizes over geometric distributions alongside momentum~\cite{momentum,si_ni_fgsm} and input diversity~\cite{inputdiversity,ti_fgsm}. We apply this transformation scheme in surface UV coordinates to optimize one world-bound texture across varied poses.

\paragraph{Attacks on embodied action models.}
Early studies establish VLA vulnerability to visual perturbations~\cite{roboticattack} and prompt attacks~\cite{robopair}. Subsequent attacks optimize 2D patches with UPA-RFAS~\cite{upa} and EDPA~\cite{edpa}, extend patches to world action models with TAPDreamer~\cite{tapdreamer}, perturb 3D object meshes via Tex3D~\cite{tex3d} and UniTexture~\cite{unitexture}, or hijack CoT reasoning with printed tablecloths in TRAP~\cite{trap}. Unlike camera patches, object meshes, or CoT surfaces (\cref{tab:related_comparison}), a persistent workspace cloth is decoupled from task objects, enters both views under evolving occlusions, and induces large-footprint renderer offsets. RefCloth addresses these constraints via shared UV diversification and same-path reference calibration using only source action-token features.

\paragraph{Feature objectives and rendering consistency.}
Intermediate feature objectives improve transfer~\cite{ila,fia,naa,ilpd}, yet optimizing 3D textures via differentiable rendering~\cite{softras,nvdiffrast,pytorch3d} introduces a synthesis gap to native simulation that biases feature baselines. While prior 3D attacks align image formation photometrically~\cite{fca,das,tex3d,unitexture}, we hold the renderer fixed and route the base cloth through the same synthesis path, canceling static path offsets at initialization.
\section{Method}
\label{sec:method}
To disrupt closed-loop control without altering objects or cameras, RefCloth learns one workspace texture on a frozen source policy (\cref{fig:method}), combining bounded UV parameterization (\cref{fig:method}b) with surface-augmented, path-calibrated optimization (\cref{fig:method}a) for frozen native transfer (\cref{fig:method}c).

\subsection{Persistent Scene Carrier}
\label{sec:problem}
We consider an attacker who controls only the appearance of a predefined background cloth whose geometry and placement remain fixed within each scene. While task objects, dynamics, camera views, instructions, and policy weights remain unchanged, the cloth naturally enters visual observations as the robot and objects evolve in closed loop.

Using gradients from a frozen source policy $\pi_s$ under bounded amplitude and variation constraints, we optimize one texture on source observations and deploy it frozen across unseen tasks and target policies $\{\pi_t\}$ in native simulation. Matched scene-state rendering and separate Original, Plain, and adversarial rollouts isolate the net texture effect.

\subsection{One Surface, Two Camera Projections}
\label{sec:texture_optimization}
\paragraph{Bounded control grid.}
Starting from a uniform base cloth $T_0\in[0,1]^{H\times W\times C}$, a coarse control grid $\theta\in\mathbb R^{K\times K\times C}$ defines the full-resolution texture
\begin{equation}
T_\theta=\operatorname{clip}_{[0,1]}\!\left(T_0+C_{\mathrm{mask}}\odot\mathcal I_{\mathrm{bary}}(\epsilon\tanh\theta)\right),
\label{eq:texture}
\end{equation}
where barycentric interpolation $\mathcal I_{\mathrm{bary}}$ uses convex weights over UV triangles and coverage mask $C_{\mathrm{mask}}\in[0,1]$ ensures $\|T_\theta-T_0\|_\infty\le\epsilon$.

\paragraph{Camera projections of the shared surface.}
For scene state $x$, shared UV transformation $u$, and camera view $v$, differentiable rasterization~\cite{nvdiffrast} gives
\begin{equation}
\widetilde I_v=M_v(x)\odot R_v(T_\theta,u;x)+(1-M_v(x))\odot I_v(x).
\label{eq:render}
\end{equation}
Because both views share $u$, each camera projects its visible slice of the shared pattern, which mask $M_v$ composites with clean frame $I_v$ under view-specific occlusions.

Feeding both composited views into the frozen source VLA with the original instruction and proprioception yields the flattened final-layer action-token features
\begin{equation}
z_\theta=\operatorname{vec}\!\bigl(h_R(x,T_\theta,u)\bigr)\in\mathbb R^m,
\label{eq:features}
\end{equation}
where $R$ denotes the differentiable synthesis path.

\subsection{Path-Calibrated Optimization with Surface Augmentation}
\label{sec:matched_reference}
When a texture is optimized through a differentiable synthesis path $R$ but deployed in a native simulator $N$, the training objective must accurately measure texture-induced changes while preventing the pattern from overfitting to a fixed surface pose. RefCloth addresses these two requirements through path-consistent reference calibration and surface-coordinate augmentation.

\paragraph{A path-consistent feature baseline.}
To measure the effect of changing the cloth pattern, we pass the unperturbed base cloth $T_0$ through the exact same differentiable renderer, compositing, and preprocessing path to obtain $z_0=\operatorname{vec}(h_R(x,T_0,u))$. We maximize the detached-reference feature displacement
\begin{equation}
\mathcal D_{\mathrm{FD}}(\theta;x,u)=\frac1m\|z_\theta-\operatorname{sg}(z_0)\|_2^2.
\label{eq:feature_displacement}
\end{equation}
With $\Delta_\theta=T_\theta-T_0$, the full training objective under surface transformation distribution $P$ is
\begin{equation}
\begin{aligned}
\mathcal J_M(\theta)={}&\lambda_f\mathbb E_{x,u\sim P}[\mathcal D_{\mathrm{FD}}(\theta;x,u)]\\
&-\lambda_a\frac{\|\Delta_\theta\|_F^2}{HWC}-\lambda_{\mathrm{tv}}\operatorname{TV}(\Delta_\theta),
\end{aligned}
\label{eq:objective}
\end{equation}
where amplitude and anisotropic spatial TV penalties regulate perturbation magnitude and smoothness via
\begin{equation}
\operatorname{TV}(\Delta)=\frac{\|D_v\Delta\|_1}{(H-1)WC}+\frac{\|D_h\Delta\|_1}{H(W-1)C},
\label{eq:tv}
\end{equation}
with $D_v,D_h$ taking adjacent-pixel differences.

\paragraph{Bias invariance and gradient calibration.}
For fixed $x,u$, let $z_N$ denote native-simulator base-cloth features, and decompose the synthesis-path discrepancy as $z_\theta = z_N(\theta) + b + r_\theta$, where $b = z_0 - z_N$ is the texture-independent baseline offset (\cref{fig:deployment}b) and $r_\theta$ is the texture-dependent residual ($r_0=0$). While the true native displacement is $d^*(\theta)=z_N(\theta)-z_N$, a cross-path native reference yields $z_\theta - z_N = d_\theta + b$ (with $d_\theta = z_\theta - z_0 = d^*(\theta) + r_\theta$), expanding as
\begin{equation}
\mathcal D_N=\mathcal D_{\mathrm{FD}}+\frac2m d_\theta^\top b+\frac1m\|b\|_2^2.
\label{eq:decomposition}
\end{equation}
Matching the reference path cancels the static offset $b$ for all $\theta$, leaving only $r_\theta$. Under common regularization and sampling, the exact gradient difference is
\begin{equation}
g_N-g_M=\frac{2\lambda_f}{m}\mathbb E_{x,u}[J_\theta^\top b],
\label{eq:gradient}
\end{equation}
where $J_\theta=\partial z_\theta/\partial\operatorname{vec}(\theta)$. At $\theta=0$, $d_0=0$ gives $\nabla_\theta\mathcal D_{\mathrm{FD}}(0)=0$, while the cross-path objective retains $\frac{2}{m}\mathbb E[J_0^\top b]$, so path matching removes zero-point drift.

\paragraph{Augmentation in surface coordinates and local geometry.}
At each step, we sample a shared UV rotation, reflection, and translation $u\sim P$ for the batch, rendering the same transformed pattern into both views (\cref{eq:render}). Because uniform $T_0$ is UV-invariant, its detached features $z_0$ are cached per observation. In a locally differentiable neighborhood of $\theta=0$ with Jacobian $A_{x,u}=\partial z_\theta(x,u)/\partial\operatorname{vec}(\theta)|_{\theta=0}$, the calibrated objective admits the quadratic approximation $\mathbb E_{x,u\sim P}[\mathcal D_{\mathrm{FD}}(\theta;x,u)]\approx\operatorname{vec}(\theta)^\top H_P\operatorname{vec}(\theta)$ with $H_P=\frac{1}{m}\mathbb E_{x,u\sim P}[A_{x,u}^\top A_{x,u}]$ (derivations in \cref{supp:theory}). Thus, the shared surface defines how dual views form $A_{x,u}$, reference matching removes the linear bias $\frac{2}{m}\mathbb E[A_{x,u}^\top b]$, and UV augmentation averages sensitivity across poses in $H_P$.

\subsection{Training and Frozen Deployment}
\label{sec:training_deployment}
We optimize control grid $\theta$ via projected sign-gradient ascent, averaging observation gradients over each mini-batch before adding texture regularizers once. Training and export share the full-resolution composition in \cref{eq:texture}, with orientation conversion and 8-bit quantization applied once to export a single PNG. During evaluation, the native simulator loads this static PNG once and renders it in closed loop as the target policy executes the task without test-time adaptation or target access, while separate Plain and random controls isolate carrier placement from learned spatial structure.
\section{Experiments}
\label{sec:experiments}
RefCloth exploits a persistent workspace surface to access both robot camera views throughout closed-loop manipulation. We empirically validate this premise across four dimensions, (i) zero-shot cross-task and cross-model transfer across 40 manipulation tasks, (ii) dual-view exposure mechanics and physical action attribution, (iii) comparisons against patch and carrier baselines, and (iv) ablations on surface UV diversification and path-consistent calibration.

\begin{figure*}[t]
\centering
\includegraphics[width=\textwidth]{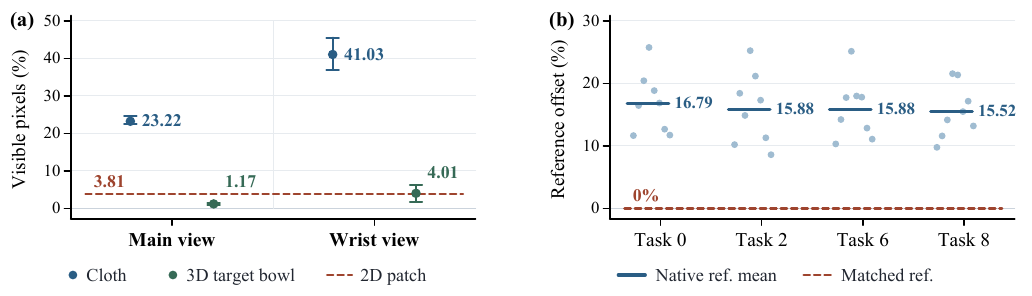}
\caption{\textbf{Carrier footprint and clean-reference offset.}
(a) Visible pixel fractions (median and IQR at \(224\times224\)) of the cloth and target bowl in both views across 32 fixed LIBERO-Spatial states, compared with the nominal footprint of a \(50\times50\) UPA-RFAS patch on a \(256\times256\) frame (dashed line).
(b) Relative Plain feature offset \(100\lVert z_R^0-z_{\mathrm{ref}}\rVert_2/\lVert z_N^0\rVert_2\) across the 32 Spatial states, where \(z_R^0\) and \(z_N^0\) denote differentiable- and native-path Plain features. Blue dots and lines show states and task means under \(z_{\mathrm{ref}}=z_N^0\), while red lines show the zero offset under \(z_{\mathrm{ref}}=z_R^0\).}
\label{fig:deployment}
\end{figure*}

\subsection{Experimental Setup}
\label{sec:exp_setup}
\paragraph{Policies and tasks.}
The source policy is OpenVLA-OFT~\cite{oft} fine-tuned on LIBERO-Spatial (150K-step checkpoint). Targets are the RLinf~\cite{rlinf} full-shot SFT versions of $\pi_0$~\cite{pi0} and $\pi_{0.5}$~\cite{pi05}. We evaluate LIBERO-Spatial, Object, Goal, and Long~\cite{libero}, using ten tasks per suite and initial states 10--19 (100 episodes per policy--suite condition). Texture optimization uses source gradients only.

\paragraph{Training and deployment.}
The fixed $60\times40$\,cm non-colliding cloth has base RGB $(204,199,194)$. We interpolate a $33\times33$ control grid to a $256\times256$ texture within bound $\epsilon=64/255$ on 296 source frames from Spatial tasks $\{0,2,6,8\}$. UV augmentation~\cite{eot} samples rotations within $\pm 180^\circ$, reflections with probability $0.5$, and translations within $\pm0.04$. All main rollouts deploy the same texture in MuJoCo 3.3.0~\cite{mujoco,robosuite} (\cref{supp:implementation}).

\paragraph{Metrics and inference.}
We pair Original (no cloth), Plain (unoptimized cloth), and adversarial conditions by task and initial state. With success rates $O, P, A$, we distinguish
\begin{equation}
D=O-P,\quad G=P-A,\quad O-A=D+G,
\label{eq:metrics}
\end{equation}
where carrier effect $D$ and net texture effect $G$ are measured in percentage points (pp, larger $G$ indicates a stronger attack). We report suite outcomes and four-suite means under independent policy noise, using exact paired McNemar tests (paired by task and initial state) with Holm correction across each eight-condition family.

\subsection{Main Results: Cross-Task and Cross-Model Transfer}
\label{sec:mainresults}
Following \cref{sec:training_deployment}, we deploy one frozen export across tasks and policies (\cref{tab:main_transfer}). Adding Plain changes success by at most 8\,pp in seven of eight conditions, while our net metric $G=P-A$ explicitly isolates learned texture effects from the larger carrier shift on $\pi_0$-Long. On the same carrier, RefCloth lowers success by a further 28--77\,pp across all eight conditions, averaging 62.50\,pp on $\pi_0$ and 43.75\,pp on $\pi_{0.5}$. Object shows large reductions on both policies (58 and 62\,pp), confirming that one texture learned on tabletop Spatial scenes transfers to the floor support plane where Object stages its targets.

\begin{table}[t]
\centering\small
\setlength{\tabcolsep}{4.5pt}
\caption{\textbf{One frozen texture across tasks and policies.} Success rates (\%) and net texture reduction $G=P-A$ (pp) over 100 episodes per suite. Original and Plain separate carrier placement from texture optimization.}
\label{tab:main_transfer}
\begin{tabular}{@{}lrrrr@{}}
\toprule
Suite & Original & Plain & RefCloth & $G\!\uparrow$\\
\midrule
\multicolumn{5}{c}{$\pi_0$}\\
Spatial & 94 & 89 & 12 & 77\\
Object & 96 & 88 & 30 & 58\\
Goal & 90 & 83 & 9 & 74\\
Long & 79 & 43 & 2 & 41\\
Mean & 89.75 & 75.75 & 13.25 & \textbf{62.50}\\
\midrule
\multicolumn{5}{c}{$\pi_{0.5}$}\\
Spatial & 98 & 98 & 60 & 38\\
Object & 99 & 100 & 38 & 62\\
Goal & 97 & 95 & 67 & 28\\
Long & 82 & 78 & 31 & 47\\
Mean & 94.00 & 92.75 & 49.00 & \textbf{43.75}\\
\bottomrule
\end{tabular}
\end{table}

\begin{table}[t]
\centering\small
\setlength{\tabcolsep}{5pt}
\caption{\textbf{Matched-statistics random controls.} Four-suite mean success rate $A$ (\%) / net drop $G$ (pp) over 800 episodes per texture.}
\label{tab:matched_random}
\begin{tabular}{@{}lrr@{}}
\toprule
Texture & $\pi_0$: $A/G$ & $\pi_{0.5}$: $A/G$\\
\midrule
Random 101 & 57.75/18.00 & 87.75/5.00\\
Random 102 & 53.50/22.25 & 85.00/7.75\\
Random 103 & 56.75/19.00 & 86.50/6.25\\
Random mean & 56.00/19.75 & 86.42/6.33\\
RefCloth & \textbf{13.25/62.50} & \textbf{49.00/43.75}\\
\bottomrule
\end{tabular}
\end{table}

\paragraph{Transfer beyond texture-training tasks.}
On the six Spatial tasks excluded from the source pool, RefCloth yields mean net reductions of 70\,pp on $\pi_0$ and 35\,pp on $\pi_{0.5}$, lowering success on 37/40 ($\pi_0$) and 31/40 ($\pi_{0.5}$) tasks across all suites to confirm unseen-task transfer.

\paragraph{Learned structure versus random patterning.}
We compare RefCloth with three random cloths matched in RMS and maximum absolute change, with TV within 3.8\% (\cref{tab:matched_random}). RefCloth exceeds the mean random control by 42.75\,pp on $\pi_0$ and 37.42\,pp on $\pi_{0.5}$, with all 24 paired contrasts surviving Holm correction ($p_H\le 0.0162$) alongside an attenuated lower-RMS Object control (+21\,pp over random at $\mathrm{RMS}\approx 24$, $p_H=5.83\times 10^{-5}$). Transfer thus depends on learned spatial structure rather than matched amplitude and TV statistics (see \cref{sec:robotwin_supp} for X-VLA~\cite{xvla} and RoboTwin~2.0~\cite{robotwin}).

\subsection{Dual-View Exposure and Policy Response}
\label{sec:camera_response}
The two projections in \cref{eq:render} expose different parts of the shared surface. We measure input visibility, test closed-loop camera interventions on the target policy, and probe source responses (protocols in \cref{supp:motivation}).

\begin{figure*}[t]
\centering
\includegraphics[width=\textwidth]{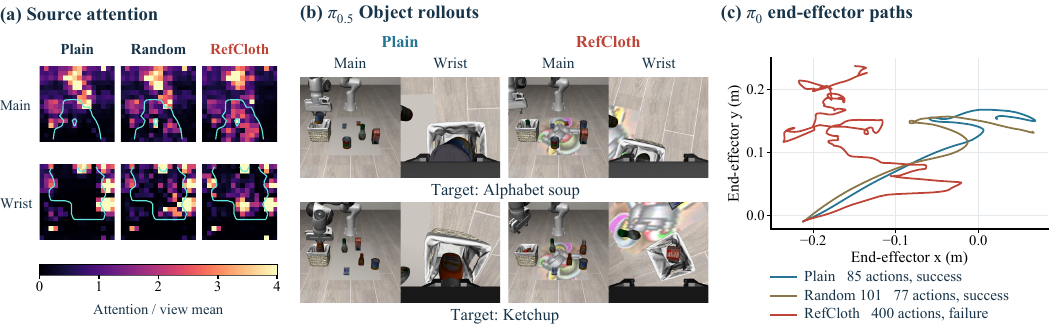}
\caption{\textbf{Visual response and closed-loop behavior.} (a) Final-layer source action-token attention over $16\times16$ visual patches at the first fixed state (Spatial task 0, initial state 21, step 10). Color is normalized by the within-view mean and clipped at four for display. Cyan contours show marker-derived cloth support. Random uses texture 101. (b) Full main and wrist frames from late Plain and RefCloth rollouts on two $\pi_{0.5}$ Object tasks at initial state 10. Plain succeeds. RefCloth places a green bottle and a milk carton in the basket, respectively. (c) Recorded XY end-effector paths on $\pi_0$-Spatial task 0, initial state 10. The numbers give executed actions.}
\label{fig:camera_behavior}
\end{figure*}

\paragraph{Surface exposure across suites.}
Compared with a target bowl or nominal 2D patch, the cloth covers a much larger pixel fraction in both views (\cref{fig:deployment}a) and exhibits larger wrist-view support across all target--suite rollouts (\cref{tab:main_exposure}). Even in Object where floor placement limits main-view coverage to 11.9\%, the moving wrist view exposes 34.4--35.7\% of the image during manipulation, showing that third-person coverage alone understates surface exposure.

\begin{table}[t]
\centering\small
\setlength{\tabcolsep}{5pt}
\caption{\textbf{Dual-view cloth support across suites.} Percentage of native pixels where Plain and cloth-hidden renders differ by $>10$ RGB levels (median over 30 episode medians).}
\label{tab:main_exposure}
\begin{tabular}{@{}lrrrr@{}}
\toprule
 & \multicolumn{2}{c}{$\pi_0$} & \multicolumn{2}{c}{$\pi_{0.5}$}\\
\cmidrule(lr){2-3}\cmidrule(l){4-5}
Suite & Main & Wrist & Main & Wrist\\
\midrule
Spatial & 23.5 & 41.3 & 23.7 & 39.7\\
Object & 11.9 & 35.7 & 11.9 & 34.4\\
Goal & 22.4 & 35.6 & 22.4 & 37.9\\
Long & 17.9 & 33.4 & 18.7 & 22.8\\
\bottomrule
\end{tabular}
\end{table}

\begin{table}[t]
\centering\footnotesize
\setlength{\tabcolsep}{3.2pt}
\caption{\textbf{Closed-loop camera interventions on $\pi_{0.5}$-Object.} 100 paired episodes per condition under synchronized noise, showing success (\%), net drop $G$ (pp), mean query steps, and Holm-adjusted McNemar $p_H$ vs.\ Plain.}
\label{tab:camera_intervention}
\begin{tabular}{@{}lrrrr@{}}
\toprule
Condition & Succ.\ (\%) & $G\!\uparrow$ (pp) & Steps & $p_H$ vs.\ Plain\\
\midrule
Plain & 98 & 0 & 32.4 & ---\\
Main-only (M) & 88 & 10 & 36.8 & .00885\\
Wrist-only (W) & 75 & 23 & 44.3 & $4.7\times 10^{-6}$\\
Both (M+W) & \textbf{34} & \textbf{64} & 66.1 & $<10^{-18}$\\
\bottomrule
\end{tabular}
\end{table}

\paragraph{Texture exposure and target closed-loop camera interventions.}
At matched Object camera states, RefCloth alters roughly three times as many preprocessed wrist pixels as main-view pixels, with over twice the pixel RMS (\cref{fig:motivation}a), lowering paired closed-loop success by 50.0\,pp on $\pi_0$ and 66.7\,pp on $\pi_{0.5}$. To test how target failures depend on each camera in closed loop, we evaluate 100 paired $\pi_{0.5}$-Object episodes per condition under synchronized query noise, rendering Plain or RefCloth at the exact camera exposure state along each rollout (\cref{fig:motivation}b, \cref{tab:camera_intervention}). Perturbing only the wrist view causes over twice the success drop of perturbing only the main view ($G=23$ vs.\ $10$\,pp, $+13$\,pp difference, $p_H=0.00885$). Moreover, simultaneous exposure in both views ($G=64$\,pp) exceeds the sum of single-view drops by $31$\,pp, as wrist perturbation lowers success by $23$\,pp when the main view is Plain, but by $54$\,pp when the main view is also RefCloth ($p_H=7.91\times 10^{-15}$), with 38/98 Plain-successful episodes surviving both single-view attacks individually yet failing under Both. Because the floor-placed cloth has a larger footprint and RMS in the wrist view (\cref{fig:motivation}a), these results reflect the combined effect of view-specific surface exposure and camera perspective.

\paragraph{Source-policy camera attribution and rollout behavior.}
To probe how each view affects immediate source actions, we substitute RefCloth into each camera at eight fixed OpenVLA-OFT states. Let $a_{PP}, a_{AP}, a_{PA}, a_{AA}$ be the first five normalized pose actions under the four Main/Wrist combinations ($P$ Plain, $A$ RefCloth), with shift $\Delta a = a_{AA} - a_{PP}$. The symmetric marginal contributions
\begin{equation}
\begin{aligned}
\phi_m&=\tfrac12[(a_{AP}-a_{PP})+(a_{AA}-a_{PA})],\\
\phi_w&=\tfrac12[(a_{PA}-a_{PP})+(a_{AA}-a_{AP})]
\end{aligned}
\label{eq:view_attribution}
\end{equation}
satisfy $\phi_m+\phi_w=\Delta a$. The signed wrist projection share $\langle\phi_w,\Delta a\rangle/\|\Delta a\|_2^2$ averages 85.0\% at step 10 and 40.2\% at step 50 across four tasks. Furthermore, decomposing the source feature-displacement gradient $\nabla_\theta\mathcal D_{\mathrm{FD}} = g_m + g_w$ across cameras at steps 0, 1,875, 3,750, and 5,000 across four training seeds (32 batches per checkpoint) shows that main and wrist gradient norms remain comparable throughout optimization, with near-orthogonal directions (median cosine $0.005$--$0.054$) and median wrist projection shares $\langle g_w, g_m+g_w\rangle/\|g_m+g_w\|_2^2$ of $50.4\%$--$57.2\%$. Consistent with these signals, source action-token attention over the cloth exceeds the mean random control at 8/8 main-view and 7/8 wrist-view states (\cref{fig:camera_behavior}a), while target rollouts exhibit distractor grasping on $\pi_{0.5}$ (\cref{fig:camera_behavior}b) and prolonged uncompleted trajectories on $\pi_0$ (\cref{fig:camera_behavior}c).

\subsection{Comparison with Existing Attacks}
\label{sec:baselines}
We compare optimization objectives on the shared cloth carrier and on each method's native attack surface.

\begin{table}[t]
\centering\small
\setlength{\tabcolsep}{2.3pt}
\caption{\textbf{Cloth-carrier transfer.} Success $A$ (\%) / net drop $G$ (pp) over 100 episodes per suite (S/O/G/L: Spatial/Object/Goal/Long). Bold marks the strongest attack per suite.}
\label{tab:adapted}
\begin{tabular}{@{}lrrrrr@{}}
\toprule
Method & S & O & G & L & Mean\\
\midrule
\multicolumn{6}{c}{$\pi_0$}\\
Native ref. & \textbf{9/80} & 28/60 & 21/62 & 4/39 & 15.50/60.25\\
EDPA & 23/66 & \textbf{14/74} & 18/65 & \textbf{2/41} & 14.25/61.50\\
UPA-RFAS & 27/62 & 35/53 & 32/51 & 7/36 & 25.25/50.50\\
RefCloth & 12/77 & 30/58 & \textbf{9/74} & \textbf{2/41} & \textbf{13.25/62.50}\\
\midrule
\multicolumn{6}{c}{$\pi_{0.5}$}\\
Native ref. & 67/31 & 80/20 & 66/29 & 33/45 & 61.50/31.25\\
EDPA & 77/21 & 57/43 & \textbf{59/36} & 35/43 & 57.00/35.75\\
UPA-RFAS & 83/15 & 55/45 & 68/27 & 51/27 & 64.25/28.50\\
RefCloth & \textbf{60/38} & \textbf{38/62} & 67/28 & \textbf{31/47} & \textbf{49.00/43.75}\\
\bottomrule
\end{tabular}
\end{table}

\begin{table}[t]
\centering\small
\setlength{\tabcolsep}{4pt}
\caption{\textbf{Native-surface deployment.} Successful episodes out of 400 and success rate (\%) across four suites under each method's native surface and training configuration.}
\label{tab:native_forms}
\begin{tabular}{@{}llrr@{}}
\toprule
Method & Surface & $\pi_0$ & $\pi_{0.5}$\\
\midrule
Original & None & 359 (89.75) & 376 (94.00)\\
Random & Patches & 290 (72.50) & 346 (86.50)\\
EDPA & Patches & 197 (49.25) & 273 (68.25)\\
UPA-RFAS & Patch & 253 (63.25) & 334 (83.50)\\
RefCloth & Cloth & \textbf{53 (13.25)} & \textbf{196 (49.00)}\\
\bottomrule
\end{tabular}
\end{table}

\paragraph{Common background carrier.}
We adapt EDPA~\cite{edpa} and UPA-RFAS~\cite{upa} to the RefCloth carrier under identical source pools and update budgets (\cref{tab:adapted}). On $\pi_0$, where surface-augmented optimization already drives task success near the floor (13.25\%), RefCloth achieves the strongest overall disruption ($G=62.50$\,pp), outperforming UPA-RFAS by 12.00\,pp and edging out EDPA (61.50\,pp). On the more resilient $\pi_{0.5}$ policy, RefCloth establishes a +8.00\,pp mean advantage over EDPA (43.75 vs.\ 35.75\,pp, with significant paired gains on Spatial and Object, $p_H=0.038$) and +15.25\,pp over UPA-RFAS (28.50\,pp, higher across all eight conditions). Across four independent training seeds on $\pi_{0.5}$ (40 tasks per seed), RefCloth achieves a four-suite mean $\bar G$ of $39.56\pm 6.86$\,pp, outperforming both cloth-adapted EDPA ($28.38\pm 3.56$\,pp, $+11.19$\,pp mean gain) and Native reference ($27.62\pm 2.03$\,pp, $+11.94$\,pp mean gain). Directly optimizing action-token feature displacement thus provides stronger transferability to robust continuous-control policies.

\paragraph{Original attack surfaces.}
When EDPA and UPA-RFAS use their native camera-coordinate patches and protocols, RefCloth lowers mean success by 36.00\,pp on $\pi_0$ and 19.25\,pp on $\pi_{0.5}$ relative to EDPA (\cref{tab:native_forms}). Compared with 2D patches constrained by localized footprints (\cref{fig:deployment}a) and viewpoint sensitivity, a persistent 3D surface carrier yields much stronger closed-loop disruption.

\subsection{Surface Augmentation Across Source Pools}
\label{sec:task_scaling}
We test how UV diversification interacts with source-pool coverage (one vs.\ four Spatial training tasks, \cref{tab:task_uv}).

\begin{table}[t]
\centering\small\setlength{\tabcolsep}{3pt}
\caption{\textbf{Source-pool size and UV augmentation.} Cells are successes out of 100 / $G$ (pp). Plain counts are in \cref{tab:main_transfer}.}
\label{tab:task_uv}
\begin{tabular}{@{}llrrrr@{}}\toprule
Tasks & UV & Spatial & Object & Goal & Long\\\midrule
\multicolumn{6}{c}{$\pi_0$}\\
One & Off & 49/40 & 65/23 & 63/20 & 19/24\\
One & On & 14/75 & 6/82 & 17/66 & 1/42\\
Four & Off & 45/44 & 61/27 & 40/43 & 13/30\\
Four & On & 12/77 & 30/58 & 9/74 & 2/41\\
\midrule
\multicolumn{6}{c}{$\pi_{0.5}$}\\
One & Off & 97/1 & 76/24 & 82/13 & 65/13\\
One & On & 44/54 & 61/39 & 50/45 & 39/39\\
Four & Off & 96/2 & 88/12 & 84/11 & 54/24\\
Four & On & 60/38 & 38/62 & 67/28 & 31/47\\
\bottomrule\end{tabular}\end{table}

\paragraph{UV augmentation and held-out transfer.}
Holding the reference fixed, enabling UV augmentation increases $G$ across all sixteen suite--model conditions ($p_H\le0.00455$), raising four-suite mean $G$ by 26.50\,pp on $\pi_0$ and 31.50\,pp on $\pi_{0.5}$ for the four-task pool. Notably, with UV enabled, the one-task pool already achieves strong four-suite mean reductions (66.25\,pp on $\pi_0$ and 44.25\,pp on $\pi_{0.5}$), while the four-task pool redistributes gains toward suites such as $\pi_0$-Goal and $\pi_{0.5}$-Object/Long. On the six Spatial tasks excluded from both pools (60 episodes per target), UV augmentation reduces success counts by 25 and 33 on $\pi_0$ and $\pi_{0.5}$ for the one-task pool, and by 15 and 19 for the four-task pool, confirming that surface-coordinate diversity drives unseen-task reuse even from a single source task.

\subsection{Path Calibration and Update Granularity}
\label{sec:reference_analysis}
We evaluate how removing the static clean-path offset $b$ shapes gradients, feature displacement, and transfer.

\begin{figure}[t]
\centering
\includegraphics[width=\columnwidth]{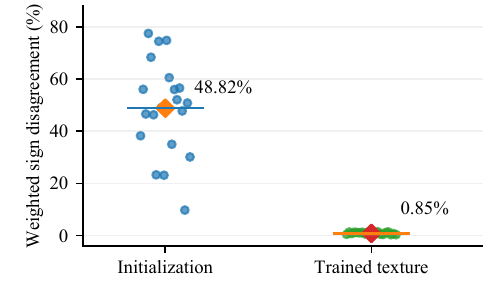}
\caption{\textbf{Reference-dependent update directions.} Gradient-weighted sign disagreement for 20 paired batches of the full objective. Points show batches; diamonds show means.}
\label{fig:full_gradients}
\end{figure}

\paragraph{Gradient alignment and displacement fidelity.}
Sign disagreement between native and matched gradients caused by $J_\theta^\top b$ (\cref{eq:gradient}) accounts for 48.82\% of the matched gradient magnitude at initialization and drops to 0.85\% after training (\cref{fig:full_gradients}), closely matching the first-order prediction. At the trained texture across the 32 Spatial states from \cref{fig:deployment}b, the matched displacement $d_M$ maintains strong positive directional alignment with the true native displacement $d^*$ (mean cosine $0.864$, with relative residual $\|r_\theta\|_2/\|d^*\|_2 = 0.490$). Path consistency thus anchors optimization by eliminating the static zero-point gradient drift that otherwise misguides early updates near the base cloth.

\paragraph{Closed-loop transfer, schedule switching, and update granularity.}
In closed-loop evaluation, while $\pi_0$ is already driven near its performance floor under either reference (+2.25\,pp gain), path matching boosts four-suite mean $G$ on the more resilient $\pi_{0.5}$ policy by +12.50\,pp in the main benchmark (\cref{fig:motivation}c, \cref{tab:adapted}) and from $27.62\pm 2.03$ to $39.56\pm 6.86$\,pp across four seeds. Testing step-500 reference-switching schedules across three seeds on $\pi_{0.5}$-Spatial/Object yields a monotonic progression in two-suite mean $G$, from $34.33\pm 9.28$\,pp (full Native) to $38.00\pm 12.32$\,pp (first-500 Matched$\to$Native) and $40.67\pm 6.60$\,pp (first-500 Native$\to$Matched), reaching $48.50\pm 3.97$\,pp under full Matched. Sustaining path consistency throughout optimization thus maximizes disruption and reduces cross-seed variance. At an equal 40,000-visit budget on $\pi_{0.5}$-Spatial/Object, full matching also improves two-suite mean $G$ by 30.00\,pp at batch one ($p\le1.31\times10^{-7}$) and 24.50\,pp at batch eight.
\section{Conclusion and Discussion}
\label{sec:conclusion}
Assessing the practical visual vulnerability of closed-loop VLA policies requires moving beyond sensor-hijacking patches and object-specific meshes toward low-privilege scene perturbations that transfer without retraining. In this work, we introduced RefCloth to show that a passive workspace tablecloth fulfills both criteria, projecting one world-bound texture consistently into static third-person and moving wrist views while leaving manipulation objects untouched. By combining shared surface UV transformations with a same-path clean reference baseline on a frozen source policy, RefCloth generalizes across partially occluded surface slices and eliminates static cross-renderer drift at initialization. Across 40 LIBERO tasks, a single frozen texture transfers zero-shot from OpenVLA-OFT to $\pi_0$ and $\pi_{0.5}$ far beyond statistics-matched random controls, while closed-loop camera interventions reveal that moving wrist views dominate standalone disruption and act synergistically with the third-person view. Extending this framework to physical cloth prints under real-world lighting, fabric deformation, and camera calibration offers a promising direction for auditing and hardening embodied robot policies.

{\small
\bibliographystyle{ieeenat_fullname}
\nocite{florence2,sapien,curobo}
\bibliography{main}}

\clearpage
\maketitlesupplementary
\appendix
\setcounter{page}{1}
\setcounter{figure}{0}\setcounter{table}{0}\setcounter{equation}{0}
\renewcommand{\thepage}{S\arabic{page}}
\renewcommand{\thefigure}{S\arabic{figure}}
\renewcommand{\thetable}{S\arabic{table}}
\renewcommand{\theequation}{S\arabic{equation}}
\renewcommand{\theHequation}{supp.\arabic{equation}}
The supplement provides complete implementation and evaluation protocols, reports extended cross-task, cross-backbone, and cross-simulator benchmarks alongside dual-view diagnostic audits, and formalizes why same-path reference matching eliminates static rendering drift during source optimization.

\section{Implementation and Evaluation Details}
\label{supp:implementation}

To ensure that every comparison in the main paper isolates texture optimization from confounding implementation differences, we first specify the exact checkpoints, rendering pipelines, carrier geometries, and baseline training budgets used across all experiments.

\paragraph{Policy checkpoints.}
We adopt the 150K-step
\href{https://huggingface.co/moojink/openvla-7b-oft-finetuned-libero-spatial}{OpenVLA-OFT LIBERO-Spatial checkpoint}
as our frozen source policy, and evaluate zero-shot transfer against the RLinf
\href{https://huggingface.co/RLinf/RLinf-Pi0-LIBERO-130-fullshot-SFT}{$\pi_0$ full-shot SFT}
and
\href{https://huggingface.co/RLinf/RLinf-Pi05-LIBERO-130-fullshot-SFT}{$\pi_{0.5}$ full-shot SFT}
target policies. During closed-loop rollouts, both target models execute ten denoising steps per query, replan every five actions, perform ten initialization steps, and run up to a horizon of 400 action steps, receiving native $256\times256$ frames resized to $224\times224$.

\paragraph{Texture optimization and deployment.}
Our training objective weights feature displacement, squared amplitude, and anisotropic spatial total variation by $(\lambda_f,\lambda_a,\lambda_{\mathrm{tv}})=(10,0.1,0.1)$ under the hyper-parameters listed in \cref{tab:supp_baseline_config}. At each update step, a single UV transformation is sampled and shared across both camera views and all mini-batch frames to enforce multi-view geometric consistency, with regularizers evaluated on normalized $[0,1]$ textures. Optimization and PNG export invoke the exact same full-resolution composition operator, applying orientation conversion and 8-bit channel quantization once upon export.

To keep physical placement consistent across manipulation layouts, we align the non-colliding cloth carrier to the primary support plane of each suite. Because LIBERO-Object stages all movable items and the destination basket directly on the floor without a table mesh, the cloth rests on the floor support plane near the objects' centroid ($0.005\,\mathrm{m}$ above the ground plane), whereas Spatial, Goal, and Long place it on the tabletop surface, with geometry and pose held strictly static during evaluation.

\paragraph{Source observation pool.}
We construct the source training pool by executing rollouts from initial states 0 and 1 across the four Spatial training tasks and sampling one frame every ten steps (up to 40 frames per rollout). While cloth placement is jittered during rollout collection to avoid overfitting to a single static coordinate, it remains fixed throughout gradient optimization, where mini-batches are drawn uniformly with replacement. Native camera-patch baselines are trained on a matching frame schedule collected without the cloth carrier.

\paragraph{Training recipe and evaluation scope.}
The primary transfer benchmark evaluates a single frozen PNG exported from source optimization across all four suites. Multi-seed reference comparisons likewise span all four $\pi_{0.5}$ suites, whereas focused diagnostic studies examine Spatial and Object so that both tabletop and floor placements are represented. Unless noted otherwise, statistical significance is evaluated via two-sided exact paired McNemar tests with Holm--Bonferroni family-wise error correction across the eight policy--suite conditions.

\paragraph{Shared-carrier configurations.}
When benchmarking alternative objectives on the shared cloth carrier, all methods start from the identical initial parameters and process the exact same observation pool and UV sampling sequence. Outer update counts and mini-batch sizes in \cref{tab:supp_baseline_config} equalize total observation visits across competing formulations.

\begin{table*}[t]
\centering\small
\setlength{\tabcolsep}{4.5pt}
\caption{\textbf{Training configurations across methods.} Cloth rows share a common observation pool, perturbation budget $\epsilon=64/255$, Gaussian initialization $\mathcal N(0,0.01^2 I)$, and parameter bounds $[-5,5]$. ``Sign'' denotes sign-gradient ascent. Native EDPA settings apply independently to each camera patch.}
\label{tab:supp_baseline_config}
\begin{tabular}{@{}llllrrl@{}}
\toprule
Method & Surface & Source & Training pool & Updates & Batch & Optimizer / step\\
\midrule
RefCloth & Cloth & OpenVLA-OFT & 296 frames & 5,000 & 8 & Sign / 0.05\\
Native reference & Cloth & OpenVLA-OFT & Same pool & 5,000 & 8 & Sign / 0.05\\
EDPA & Cloth & OpenVLA-OFT & Same pool & 5,000 & 8 & Sign / 0.05\\
UPA-RFAS & Cloth & OpenVLA-OFT & Same pool & 5,000 & 8 & Sign / 0.05\\
\midrule
EDPA & Two $57\times57$ patches & OpenVLA-OFT & 303 no-cloth frames & 1,200 & 2 & Sign / $2/255$\\
UPA-RFAS & One $50\times50$ patch & OpenVLA & Spatial RLDS & 2,000 & 2 & AdamW / $10^{-3}$\\
Tex3D & Object texture & OFT branch & 10 frames / task & 5,000 & 10 & Upstream update\\
\bottomrule
\end{tabular}
\end{table*}

\paragraph{Cloth-adapted EDPA.}
For the cloth-adapted EDPA baseline, the visual representation concatenates 512 projected tokens from both views alongside textual input embeddings, with clean reference features generated through the identical differentiable renderer on the Plain cloth. Its objective linearly combines token-level contrastive loss and vision--language alignment with weights $1-\alpha$ and $\alpha$ ($\alpha=0.2$, temperature $0.07$), where the alignment term tracks absolute drift in visual--text cosine similarity over instruction tokens. Loss components are normalized via detached exponential moving averages (decay $0.9$, 100-update warmup) alongside squared amplitude and spatial TV penalties of weight $0.1$, while an amplitude-variant baseline replaces spatial TV with an elementwise $\ell_1$ penalty.

\paragraph{Cloth-adapted UPA-RFAS.}
\label{supp:cloth_upa}
Our cloth-adapted UPA-RFAS baseline combines bidirectional contrastive optimization (temperature $0.07$, weight $1.0$), feature-$\ell_1$ displacement (weight $0.1$), and text-steering cross-entropy (weight $0.5$) over 33 instruction-token positions excluding proprioception. Its inner robust loop runs ten steps with an auxiliary $2/255$ bound outside the cloth mask, omitting the original attention term due to OpenVLA-OFT's distinct attention interface while retaining squared and elementwise amplitude regularizers of weight $0.1$.

\paragraph{Native EDPA patches.}
For native 2D patch evaluation, we run the official EDPA release with mixture weight $0.2$, uniform random initialization, and $[0,1]$ clipping, sampling patch locations dynamically per frame during training. The unoptimized control draws two static RGB patches from $\mathcal U(0,1)$, and evaluation pairs learned and random patches at identical per-episode coordinates.

\paragraph{Native UPA-RFAS patch.}
We similarly deploy the official UPA-RFAS implementation using a Spatial checkpoint, 20 warmup iterations, 50 inner optimization steps, geometric transformations, loss weights $(1.0,1.0,0.5,0.1)$ for contrastive, attention, text, and feature-$\ell_1$ terms, and an optimizer restart at step 1,400, projecting the identical patch into both views at per-episode sampled coordinates.

\paragraph{Native object texture.}
Finally, for object-bound 3D perturbation, we evaluate the official Tex3D OFT release under a $64/255$ vertex-color perturbation budget applied directly to target object meshes.

\section{Additional Transfer Comparisons}
\label{supp:additional_results}

Building on the aggregate transfer rates reported in the main paper, this section provides complete episode-level counts across all 40 LIBERO tasks, paired statistical contrasts against existing objectives and statistics-matched random patterns, multi-seed stability analyses, and extended evaluations on a heterogeneous vision backbone and a bimanual simulator.

\Cref{tab:supp_absolute,tab:supp_native_absolute} break down raw episode success counts out of 100 per suite for both shared-cloth formulations and native camera-patch deployments.

\begin{table*}[!t]
\centering\small
\setlength{\tabcolsep}{5.5pt}
\caption{\textbf{Complete cloth-carrier performance across suites.} Successful episodes out of 100. The lower-amplitude random cloth has RMS 24.011, whereas main RefCloth has RMS 37.499, and calibrated amplitude-matched controls appear in \cref{tab:supp_matched_random}.}
\label{tab:supp_absolute}
\begin{tabular}{@{}llrrrrrrr@{}}
\toprule
Target & Suite & Original & Plain & Random cloth & Native ref. & EDPA & UPA-RFAS & RefCloth\\
\midrule
$\pi_0$ & Spatial & 94 & 89 & 62 & 9 & 23 & 27 & 12\\
 & Object & 96 & 88 & 73 & 28 & 14 & 35 & 30\\
 & Goal & 90 & 83 & 79 & 21 & 18 & 32 & 9\\
 & Long & 79 & 43 & 36 & 4 & 2 & 7 & 2\\
\midrule
$\pi_{0.5}$ & Spatial & 98 & 98 & 97 & 67 & 77 & 83 & 60\\
 & Object & 99 & 100 & 92 & 80 & 57 & 55 & 38\\
 & Goal & 97 & 95 & 90 & 66 & 59 & 68 & 67\\
 & Long & 82 & 78 & 67 & 33 & 35 & 51 & 31\\
\bottomrule
\end{tabular}
\end{table*}

\begin{table}[t]
\centering\small
\setlength{\tabcolsep}{3.5pt}
\caption{\textbf{Native-patch outcomes across suites.} Successful episodes out of 100, and totals out of 400. Original denotes the baseline unperturbed scene. Random and optimized EDPA patches share recorded placement sequences.}
\label{tab:supp_native_absolute}
\begin{tabular}{@{}lrrrr@{}}
\toprule
Suite & Original & Random & EDPA & UPA-RFAS\\
\midrule
\multicolumn{5}{c}{$\pi_0$}\\
Spatial & 94 & 87 & 59 & 74\\
Object & 96 & 81 & 62 & 74\\
Goal & 90 & 74 & 50 & 61\\
Long & 79 & 48 & 26 & 44\\
Total & 359 & 290 & 197 & 253\\
\midrule
\multicolumn{5}{c}{$\pi_{0.5}$}\\
Spatial & 98 & 94 & 88 & 94\\
Object & 99 & 96 & 76 & 94\\
Goal & 97 & 86 & 72 & 82\\
Long & 82 & 70 & 37 & 64\\
Total & 376 & 346 & 273 & 334\\
\bottomrule
\end{tabular}
\end{table}

\paragraph{Paired comparison with EDPA.}
To examine whether RefCloth's advantage over cloth-adapted EDPA holds consistently at the episode level, we pair rollouts by task and initial state across all eight policy--suite conditions (\cref{tab:supp_edpa_pairwise}). RefCloth achieves statistically significant gains on $\pi_{0.5}$-Spatial and Object ($p_H=0.038$), whereas EDPA induces stronger disruption on $\pi_0$-Object ($p_H=0.020$). Although both $\pi_0$ and $\pi_{0.5}$ employ continuous flow-matching action heads, this divergence tracks target resilience, as $\pi_{0.5}$ maintains high clean stability ($92.75\%$ Plain mean and $95\%$--$100\%$ on Spatial, Object, and Goal) that responds strongly to direct action-token displacement, while the more fragile $\pi_0$ policy is heavily degraded by either objective.

\begin{table}[htbp]
\centering\small
\setlength{\tabcolsep}{4.5pt}
\caption{\textbf{Paired comparison between RefCloth and cloth-adapted EDPA.}
$\Delta G=G_{\mathrm{RefCloth}}-G_{\mathrm{EDPA}}$ (pp).
Discordant pairs count RefCloth stronger / EDPA stronger, and
$p_H$ denotes Holm-adjusted exact McNemar $p$-values.}
\label{tab:supp_edpa_pairwise}
\begin{tabular}{llrrr}
\toprule
Target & Suite & $\Delta G$ & Pairs & $p_H$\\
\midrule
$\pi_0$ & Spatial & 11 & 18/7 & .216\\
 & Object & $-16$ & 5/21 & \textbf{.020}\\
 & Goal & 9 & 13/4 & .216\\
 & Long & 0 & 2/2 & 1.000\\
\midrule
$\pi_{0.5}$ & Spatial & 17 & 26/9 & \textbf{.038}\\
 & Object & 19 & 31/12 & \textbf{.038}\\
 & Goal & $-8$ & 10/18 & .555\\
 & Long & 4 & 14/10 & 1.000\\
\bottomrule
\end{tabular}
\end{table}

\paragraph{Cloth UPA-RFAS comparison.}
Extending the same paired analysis to cloth-adapted UPA-RFAS shows that RefCloth achieves a larger net drop $G$ in all eight policy--suite conditions (\cref{tab:supp_upa_corrected}), with significant gains on $\pi_0$-Goal ($p_H=0.000272$) as well as $\pi_{0.5}$-Spatial ($p_H=0.00134$) and Long ($p_H=0.00493$). Direct action-token feature displacement thus transfers more reliably across multi-view workspace surfaces than contrastive text-steering losses.

\begin{table}[t]
\centering\small\setlength{\tabcolsep}{5pt}
\caption{\textbf{RefCloth versus cloth-adapted UPA-RFAS.} Pairs count RefCloth stronger / UPA-RFAS stronger. $\Delta G=G_{\mathrm{RefCloth}}-G_{\mathrm{UPA}}$, and $p_H$ uses Holm adjustment across eight tests.}
\label{tab:supp_upa_corrected}
\begin{tabular}{@{}llrrr@{}}\toprule
Target & Suite & $\Delta G$ & Pairs & $p_H$\\\midrule
$\pi_0$ & Spatial & 15 & 22/7 & .0407\\
& Object & 5 & 19/14 & .974\\
& Goal & 23 & 27/4 & .000272\\
& Long & 5 & 6/1 & .375\\\midrule
$\pi_{0.5}$ & Spatial & 23 & 30/7 & .00134\\
& Object & 17 & 29/12 & .0460\\
& Goal & 1 & 10/9 & 1.000\\
& Long & 20 & 27/7 & .00493\\\bottomrule
\end{tabular}
\end{table}

\paragraph{Random cloth and amplitude-controlled comparison.}
\label{supp:random_cloth}
To verify that the advantages of learned feature optimization and path calibration are not an artifact of large perturbation norms, we further evaluate an amplitude-attenuated RefCloth pattern alongside matched controls on $\pi_{0.5}$-Object with texture RMS held near $24$, where the unoptimized random baseline is generated by bilinearly resizing a Gaussian grid to match decoded statistics ($L_\infty=57$, TV $8.035$).

Under this tighter energy budget (\cref{tab:supp_rms_controls}), RefCloth reduces success to $71/100$ ($G=29$\,pp), outperforming both the unoptimized random cloth and the batch-8 native reference by $21$\,pp ($p_H\le 5.83\times 10^{-5}$ across the four-test Holm family) and surpassing the batch-1 native reference by $13$\,pp ($p_H=0.00195$), while remaining comparable to the $\ell_1$-penalized EDPA variant ($p_H=0.307$). Learned spatial structure and reference calibration therefore retain clear statistical separability even under constrained perturbation amplitudes.

\begin{table*}[t]
\centering\small
\setlength{\tabcolsep}{5.5pt}
\caption{\textbf{RMS-controlled comparisons on $\pi_{0.5}$-Object.} All conditions utilize the cloth carrier with texture RMS held near 24. Plain baseline achieves 100/100 successes. Pairs count RefCloth stronger / comparator stronger. Adjusted $p_H$ values use a predefined four-test Holm-corrected family comparing RefCloth against each baseline.}
\label{tab:supp_rms_controls}
\begin{tabular}{@{}llrrrl@{}}
\toprule
Method / Comparator & Configuration & Successes / 100 & $G$ (pp) & Pairs & $p_H$\\
\midrule
\textbf{RefCloth (Ours)} & Attenuated grid ($\epsilon\tanh$) & \textbf{71} & \textbf{29} & --- & ---\\
\midrule
Native reference & Batch 8, spatial TV & 92 & 8 & 21/0 & $3.81\times10^{-6}$\\
Native reference & Batch 1, spatial TV & 84 & 16 & 14/1 & .00195\\
Random cloth & Fixed, unoptimized & 92 & 8 & 23/2 & $5.83\times10^{-5}$\\
EDPA & $\ell_1$ amplitude penalty & 77 & 23 & 15/9 & .307\\
\bottomrule
\end{tabular}
\end{table*}

\subsection{Main-Amplitude Random Controls}
\label{supp:matched_random_controls}
\begin{figure}[t]
\centering
\includegraphics[width=\columnwidth]{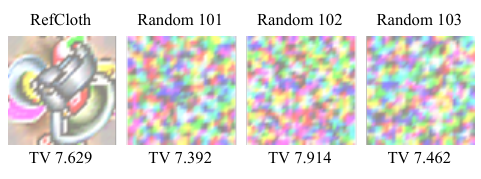}
\caption{\textbf{Exported textures in the main-amplitude random control.} Decoded 8-bit PNGs of RefCloth and three independent random controls. All patterns share RMS 37.499 and $L_\infty=64$, with calibrated TV labels indicating closely matched spatial variation.}
\label{fig:supp_matched_textures}
\end{figure}

At the primary evaluation budget ($\mathrm{RMS}=37.50$, $L_\infty=64$), we next test whether closed-loop disruption stems from learned spatial structure rather than first-order pixel contrast or spatial variation alone. Specifically, we synthesize three independent random textures matched to RefCloth in RMS ($37.50$), maximum absolute deviation ($64$), and spatial TV within $3.8\%$ (\cref{fig:supp_matched_textures}). Across all 24 paired suite--model contrasts in \cref{tab:supp_matched_random}, RefCloth outperforms every random control by wide margins, with every comparison surviving Holm correction (maximum $p_H=0.0162$).

\begin{table*}[t]
\centering\small\setlength{\tabcolsep}{6pt}
\caption{\textbf{Complete matched-statistics random comparison across all suites.} Successes out of 100. Each row compares identical task--initial-state keys. The final column reports the maximum adjusted $p$-value among the three RefCloth--random comparisons per condition.}
\label{tab:supp_matched_random}
\begin{tabular}{@{}llrrrrrr@{}}\toprule
Target & Suite & Plain & RefCloth & Random 101 & Random 102 & Random 103 & Maximum $p_H$\\\midrule
$\pi_0$ & Spatial & 89 & 12 & 65 & 62 & 65 & $4.88\times10^{-12}$\\
& Object & 88 & 30 & 67 & 64 & 68 & $9.31\times10^{-7}$\\
& Goal & 83 & 9 & 73 & 65 & 71 & $1.84\times10^{-12}$\\
& Long & 43 & 2 & 26 & 23 & 23 & $1.14\times10^{-5}$\\\midrule
$\pi_{0.5}$ & Spatial & 98 & 60 & 99 & 99 & 98 & $3.73\times10^{-10}$\\
& Object & 100 & 38 & 90 & 93 & 90 & $4.27\times10^{-14}$\\
& Goal & 95 & 67 & 90 & 81 & 90 & .0162\\
& Long & 78 & 31 & 72 & 67 & 68 & $1.14\times10^{-9}$\\\bottomrule
\end{tabular}
\end{table*}

\paragraph{Native attack surfaces and reconstruction controls.}
In native camera-coordinate attacks (\cref{tab:supp_native_absolute}), pairing learned EDPA patches with random patches at identical coordinates controls for visual occlusion, showing that learned patches outperform random occlusion in $7/8$ suites.

A parallel control is equally important for 3D object textures (Tex3D), where re-baking a mesh introduces photometric domain shift independent of adversarial optimization. Evaluating the unbaked bowl, a zero-perturbation baked bowl, and the optimized Tex3D bowl yields success counts of $148/154/151$ on $\pi_0$ and $159/159/158$ on $\pi_{0.5}$ out of 160 episodes, demonstrating that although mesh baking alters image appearance (relative image difference $0.2195$), gradient optimization adds negligible closed-loop disruption.

\begin{figure}[t]
\centering
\includegraphics[width=.94\columnwidth]{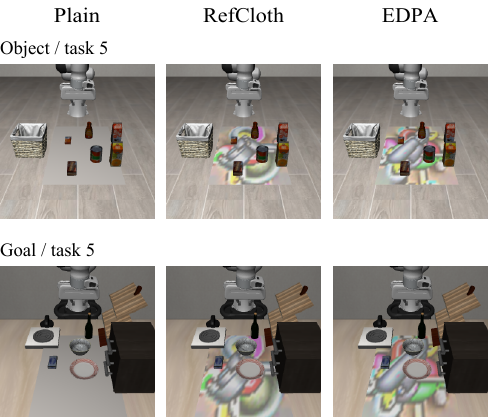}
\caption{\textbf{Additional native-rendered deployments.} Plain, RefCloth, and cloth-adapted EDPA at matched states in Object task 5 and Goal task 5. Spatial is shown in the main text, and full texture maps appear in \cref{fig:supp_texture_gallery}.}
\label{fig:suppdeployment}
\end{figure}

\begin{figure*}[t]
\centering
\includegraphics[width=\textwidth]{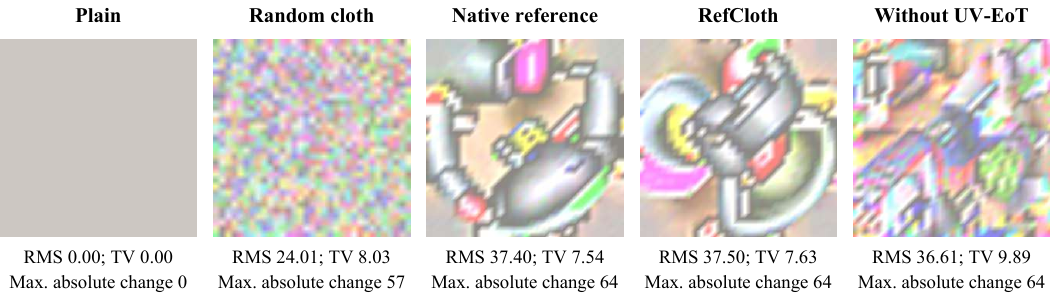}
\caption{\textbf{Gallery of deployed textures and baselines.} Exported 8-bit PNGs in UV coordinates. RMS and TV are computed directly on decoded pixel arrays. The main benchmark uses Random 101 (RMS 37.50), and the RMS 24.01 control is evaluated in \cref{tab:supp_rms_controls}.}
\label{fig:supp_texture_gallery}
\end{figure*}

\paragraph{Deployed textures.}
Visualizing the complete gallery of exported textures (\cref{fig:supp_texture_gallery}) confirms that both path-matched and native-reference variants form coherent, curved spatial contours distinct from unoptimized and random controls, whereas omitting UV augmentation yields fragmented high-TV artifacts.

\subsection{Source-Policy Evaluation}
\label{supp:source_transfer}
Before examining cross-architecture transfer, we also verify that the exported textures generalize across unseen tasks on the source architecture itself using OpenVLA-OFT's native $512\times512$ visual interface (\cref{tab:supp_source_transfer}). On the six Spatial tasks held out from the training pool, both Native and Matched textures sharply reduce success from $59/60$ (Plain) to $35/60$ and $28/60$, respectively, with no statistically significant difference between the two references ($p=0.405$), confirming that surface UV diversification drives within-architecture task generalization while same-path calibration primarily benefits cross-model transfer.

\begin{table}[tb]
\centering\small
\setlength{\tabcolsep}{4.5pt}
\caption{\textbf{Closed-loop evaluation on OpenVLA-OFT.} Successful episodes out of total under native source rendering. Held-out tasks are excluded from texture training.}
\label{tab:supp_source_transfer}
\begin{tabular}{@{}lrrr@{}}
\toprule
Spatial tasks & Plain & Native ref. & RefCloth\\
\midrule
All ten & 99/100 & 64/100 & 59/100\\
Six held out & 59/60 & 35/60 & 28/60\\
\bottomrule
\end{tabular}
\end{table}

\subsection{Multi-Seed Method Comparison and Reference Schedules}
\label{supp:reference_runs}
To ensure that our main benchmark findings do not depend on a single favorable random seed or training schedule, we next evaluate multi-seed stability, step-500 reference switching, and update granularity on $\pi_{0.5}$.

\paragraph{Four-seed comparison of Native reference, cloth-adapted EDPA, and RefCloth.}
We train Native reference, cloth-adapted EDPA, and RefCloth across four independent seeds (42, 7, 66, and 88) and evaluate each exported PNG across all four $\pi_{0.5}$ suites ($40$ tasks, $400$ episodes per seed--method condition, $4{,}800$ attack episodes total). Within every seed, all three methods share identical initial weights, observation order, and UV transformations over 5,000 updates at batch size eight, and evaluation pairs task--initial-state keys with synchronized query-indexed noise. Across seeds, cloth-adapted EDPA exhibits slightly higher decoded RMS ($39.08$--$39.29$) and spatial TV ($8.11$--$8.51$) than RefCloth ($\mathrm{RMS}=37.50$, $\mathrm{TV}=7.63$) at $L_\infty=63.99$.

As reported in \cref{tab:supp_reference_cases}, the four-seed macro-averaged net reduction strictly follows $\mathrm{RefCloth}$ ($39.56\pm 6.86$\,pp) $>$ $\mathrm{Cloth\text{-}EDPA}$ ($28.38\pm 3.56$\,pp) $>$ $\mathrm{Native}$ ($27.62\pm 2.03$\,pp). RefCloth surpasses Native reference in all $4/4$ seeds (mean gain $+11.94\pm 8.03$\,pp) and outperforms cloth-adapted EDPA in $3/4$ seeds by $+11.50$ to $+18.25$\,pp while matching it within a single episode out of 400 on Seed 88 ($-0.25$\,pp).

\begin{table}[t]
\centering\footnotesize
\setlength{\tabcolsep}{3.2pt}
\caption{\textbf{Four-seed common-carrier comparison on $\pi_{0.5}$.} Four-suite macro-averaged net reduction $\bar G$ (pp) across Spatial, Object, Goal, and Long (400 episodes per cell). $\Delta_{\mathrm{Nat}} = \bar G_{\mathrm{RefCloth}} - \bar G_{\mathrm{Native}}$, and $\Delta_{\mathrm{EDPA}} = \bar G_{\mathrm{RefCloth}} - \bar G_{\mathrm{EDPA}}$. SD is computed across the four independent training seeds.}
\label{tab:supp_reference_cases}
\begin{tabular}{@{}rrrrrr@{}}\toprule
Seed & Native & Cloth-EDPA & RefCloth & $\Delta_{\mathrm{Nat}}$ & $\Delta_{\mathrm{EDPA}}$\\\midrule
42 & 25.75 & 24.00 & \textbf{35.50} & $+9.75$ & $+11.50$\\
7 & 27.00 & 27.25 & \textbf{45.50} & $+18.50$ & $+18.25$\\
66 & 27.25 & 30.00 & \textbf{45.25} & $+18.00$ & $+15.25$\\
88 & 30.50 & \textbf{32.25} & 32.00 & $+1.50$ & $-0.25$\\
\midrule
Mean & 27.62 & 28.38 & \textbf{39.56} & $+11.94$ & $+11.19$\\
SD & 2.03 & 3.56 & 6.86 & 8.03 & 8.11\\\bottomrule
\end{tabular}
\end{table}

\paragraph{Reference-switching schedules.}
Because gradient sign disagreement between matched and native references is largest near initialization, a natural question is whether early calibration alone suffices or whether path consistency must be maintained throughout training. We test this by evaluating two 5,000-step schedules that switch references at step 500 across seeds 7, 8, and 9 on $\pi_{0.5}$-Spatial and Object ($100$ episodes per suite, $2{,}400$ episodes total, \cref{tab:supp_ref_switch}), namely \emph{Matched$\to$Native} (using $z_0$ for steps 1--500 and $z_N$ for steps 501--5,000) and \emph{Native$\to$Matched} (using $z_N$ first and $z_0$ thereafter), preserving parameters, optimizer state, and sampling order across the switch.

Across the three seeds (\cref{tab:supp_ref_switch}), two-suite mean $G$ rises monotonically from $34.33\pm 9.28$\,pp (full Native) to $38.00\pm 12.32$\,pp (Matched$\to$Native) and $40.67\pm 6.60$\,pp (Native$\to$Matched), reaching $48.50\pm 3.97$\,pp under full Matched. Calibrating only the first 500 steps gains $3.67$\,pp on average over full Native but remains $10.50$\,pp below full Matched (changing individual seeds by $+11.0$, $-7.0$, and $+7.0$\,pp), whereas switching to the matched reference after step 500 gains $6.33$\,pp over full Native yet still trails full Matched by $7.83$\,pp. Sustaining same-path calibration across the entire trajectory is therefore essential to prevent ongoing bias from the $\frac{2}{m}d_\theta^\top b$ cross-term in \cref{eq:decomposition}.

\begin{table}[t]
\centering\footnotesize
\setlength{\tabcolsep}{3.6pt}
\caption{\textbf{Step-500 reference-switching schedules on $\pi_{0.5}$-Spatial/Object.} Net reduction $G$ (pp, 100 episodes per suite, where Plain is 98/100 on Spatial and 100/100 on Object). M$\to$N and N$\to$M switch references at step 500 of 5,000 while preserving parameters, optimizer state, and sampling order. The middle block reports the two-suite mean per seed, and bottom rows report the mean and SD across seeds ($n=3$) and across all seed--suite cells ($n=6$).}
\label{tab:supp_ref_switch}
\begin{tabular}{@{}llrrrr@{}}\toprule
Seed & Scope & Native & M$\to$N & N$\to$M & Matched\\\midrule
7 & Spatial & 31.0 & 44.0 & \textbf{62.0} & 38.0\\
  & Object  & 20.0 & 29.0 & 31.0 & \textbf{62.0}\\
8 & Spatial & 29.0 & 18.0 & 28.0 & \textbf{63.0}\\
  & Object  & 38.0 & 35.0 & 39.0 & \textbf{40.0}\\
9 & Spatial & 45.0 & \textbf{62.0} & 44.0 & 38.0\\
  & Object  & 43.0 & 40.0 & 40.0 & \textbf{50.0}\\
\midrule
7 & 2-suite mean & 25.5 & 36.5 & 46.5 & \textbf{50.0}\\
8 & 2-suite mean & 33.5 & 26.5 & 33.5 & \textbf{51.5}\\
9 & 2-suite mean & 44.0 & \textbf{51.0} & 42.0 & 44.0\\
\midrule
Mean & All & 34.33 & 38.00 & 40.67 & \textbf{48.50}\\
SD & $n=3$ seeds & 9.28 & 12.32 & 6.60 & \textbf{3.97}\\
SD & $n=6$ cells & 9.46 & 14.87 & 12.03 & 11.73\\\bottomrule
\end{tabular}
\end{table}

\paragraph{Update granularity at a fixed observation budget.}
Holding the total budget fixed at 40,000 source observation visits on $\pi_{0.5}$-Spatial and Object (\cref{tab:budget}), we further compare batch size one (40,000 updates) against batch size eight (5,000 updates). Same-path reference matching improves two-suite mean $G$ by $30.00$\,pp at batch one (unadjusted exact $p\le 1.31\times 10^{-7}$ in both suites) and $24.50$\,pp at batch eight, showing that calibration benefits hold across update granularities.

\begin{table}[t]
\centering\footnotesize
\setlength{\tabcolsep}{4pt}
\caption{\textbf{Update granularity at 40,000 source visits.} Cells are successes out of 100 / $G$ (pp) on $\pi_{0.5}$-Spatial/Object (S/O), with Plain counts in \cref{tab:main_transfer}. Source visits equal outer updates times batch size, and all other settings match \cref{tab:supp_baseline_config}.}
\label{tab:budget}
\begin{tabular}{@{}rrrrrrr@{}}
\toprule
& & & \multicolumn{2}{c}{Native} & \multicolumn{2}{c}{Matched}\\
\cmidrule(lr){4-5}\cmidrule(l){6-7}
Batch & Updates & Visits & S & O & S & O\\
\midrule
8 & 5,000 & 40,000 & 67/31 & 80/20 & 60/38 & 38/62\\
1 & 40,000 & 40,000 & 85/13 & 72/28 & \textbf{54/44} & \textbf{43/57}\\
\bottomrule
\end{tabular}
\end{table}

\subsection{Cross-Backbone and Cross-Simulator Evaluation}
\label{sec:robotwin_supp}

While OpenVLA-OFT and $\pi_0/\pi_{0.5}$ operate within MuJoCo LIBERO and share SigLIP visual representations, practical deployment scenarios involve distinct vision backbones, RL-post-trained policies, and bimanual workspaces. To test generalization along all three axes, we evaluate the same $0.60 \times 0.40\,\mathrm{m}$ non-colliding cloth carrier on \textbf{X-VLA}~\cite{xvla} (built on a Florence-2 DaViT vision backbone~\cite{florence2} distinct from SigLIP and DINOv2) across all four LIBERO suites, and further deploy onto SAPIEN-based~\cite{sapien} \textbf{RoboTwin~2.0}~\cite{robotwin}.

\paragraph{Experimental setup.}
On LIBERO, we evaluate $N{=}500$ paired episodes per suite ($10$ tasks $\times$ $50$ initial states, $2{,}000$ episodes per condition) comparing \textit{Plain} ($P$), matched \textit{Random~101}, our frozen LIBERO-Spatial OpenVLA-OFT texture (\textbf{RefCloth}, \cref{tab:main_transfer}), a target-architecture self-attack (\textbf{X-VLA-Self}, $N{=}500$ per suite) trained directly on X-VLA across the four suites, and a backbone-only variant (\textbf{VLM-only}, $N{=}100$ per suite) that maximizes displacement solely on X-VLA's Florence-2 VLM representations without action-head gradients.
On RoboTwin~2.0, the cloth is placed $0.005\,\mathrm{m}$ above the workspace table and observed simultaneously by \texttt{head\_camera}, \texttt{left\_camera}, and \texttt{right\_camera} across identical paired initial states per task, comparing \textit{Original} ($O$, no cloth), \textit{Plain} ($P$), matched random controls (\textit{Random~101/102}), frozen LIBERO \textbf{RefCloth}, a RoboTwin-native texture (\textbf{RefCloth-RT}) re-optimized on SAPIEN observations of RLinf-$\pi_{0.5}$-SFT on \textit{adjust\_bottle}, and a RoboTwin-trained \textbf{X-VLA-Self} texture (all trained for $5{,}000$ updates under $\epsilon = 64/255$).
RoboTwin targets comprise the RL-finetuned \textbf{RLinf-$\pi_{0.5}$-PPO}~\cite{rlinf} (\texttt{RLinf-Pi05-RoboTwin-PPO-adjust\_bottle}, $200$ max actions, action chunk $50$, $5$ denoising steps) on \textit{adjust\_bottle} ($N{=}128$), and bimanual \textbf{X-VLA}~\cite{xvla} (horizon $300$, $10$ denoising steps, action chunk $30$, $\mathrm{cuRobo}$ planner~\cite{curobo}) across \textit{adjust\_bottle} ($N{=}128$), \textit{beat\_block\_hammer} ($N{=}100$), and \textit{place\_empty\_cup} ($N{=}100$).
Following \cref{eq:metrics}, we report carrier shift $D = O - P$, net texture drop $G = P - A$ ($\mathrm{pp}$), and two-sided exact paired McNemar $p$-values against \textit{Plain} ($p_{\text{Plain}}$) and matched random controls.

\begin{table*}[t]
\centering
\caption{\textbf{Cross-backbone evaluation on X-VLA across four LIBERO suites.} Success rates (\%) over $N{=}500$ paired episodes per suite ($2{,}000$ total, with $N{=}100$ per suite for \textit{VLM-only}). Net drop $G = P - A$ ($\mathrm{pp}$) and exact paired McNemar $p$-values ($p_{\text{Plain}}$) compare frozen OpenVLA-OFT \textit{RefCloth} (\cref{tab:main_transfer}) against \textit{Plain}. \textit{X-VLA-Self} ($N{=}500$) and \textit{VLM-only} ($N{=}100$) optimize directly against X-VLA and its Florence-2 VLM backbone, respectively, reducing four-suite mean success by $76.90$ and $78.40\,\mathrm{pp}$ ($p \le 6.89{\times}10^{-53}$ vs.\ \textit{Plain} and $p \le 1.25{\times}10^{-36}$ vs.\ \textit{RefCloth} across every suite for \textit{X-VLA-Self}).}
\label{tab:libero_xvla}
\footnotesize
\setlength{\tabcolsep}{4.8pt}
\begin{tabular}{@{}lcccccc@{}}
\toprule
Suite & Plain ($P$) & Random~101 & RefCloth (\cref{tab:main_transfer}) & $G\uparrow$ ($p_{\text{Plain}}$) & X-VLA-Self & VLM-only \\
\midrule
Spatial & 98.20 (491/500) & 97.40 (487/500) & 97.80 (489/500) & $+0.40$ (.688) & \textbf{0.40} (2/500) & 16.00 (16/100) \\
Object  & 99.60 (498/500) & 98.60 (493/500) & 89.80 (449/500) & $+9.80$ ($4.62{\times}10^{-14}$) & 12.40 (62/500) & \textbf{7.00} (7/100) \\
Goal    & 97.80 (489/500) & 97.60 (488/500) & 94.80 (474/500) & $+3.00$ (.0107) & 18.40 (92/500) & \textbf{13.00} (13/100) \\
Long    & 95.00 (475/500) & 95.20 (476/500) & 83.20 (416/500) & $+11.80$ ($1.79{\times}10^{-9}$) & 51.80 (259/500) & \textbf{41.00} (41/100) \\
\midrule
Mean    & 97.65 (1953/2000) & 97.20 (1944/2000) & 91.40 (1828/2000) & $+6.25$ & 20.75 (415/2000) & \textbf{19.25} (77/400) \\
\bottomrule
\end{tabular}
\end{table*}

\begin{table}[t]
\centering
\caption{\textbf{RoboTwin~2.0 \textit{adjust\_bottle} evaluation ($N{=}128$) on RLinf-$\pi_{0.5}$-PPO.} Net drop $G = P - A$ ($\mathrm{pp}$) is measured relative to \textit{Plain} ($82.03\%$), and $p_{\text{Plain}}$ reports two-sided exact paired McNemar tests against \textit{Plain}. \textit{X-VLA-Self} further significantly outperforms both \textit{Random~101} ($p{=}8.21{\times}10^{-4}$) and \textit{Random~102} ($p{=}4.92{\times}10^{-5}$).}
\label{tab:rt_ppo_bottle}
\footnotesize
\setlength{\tabcolsep}{3.5pt}
\begin{tabular}{@{}lcccc@{}}
\toprule
Condition & Succ. & Rate (\%) & $G\uparrow$ (pp) & $p_{\text{Plain}}$ \\
\midrule
Original ($O$)    & 118 & 92.19 & ---   & --- \\
Plain ($P$)       & 105 & 82.03 &  0.00 & --- \\
\midrule
Random~101        &  92 & 71.88 & 10.16 & .00443 \\
Random~102        &  93 & 72.66 &  9.38 & .00754 \\
\midrule
RefCloth (\cref{tab:main_transfer}) &  86 & 67.19 & 14.84 & $5.46{\times}10^{-4}$ \\
RefCloth-RT       &  84 & 65.62 & 16.41 & $4.92{\times}10^{-5}$ \\
X-VLA-Self        &  \textbf{72} & \textbf{56.25} & \textbf{25.78} & $3.61{\times}10^{-8}$ \\
\bottomrule
\end{tabular}
\end{table}

\begin{table*}[t]
\centering
\caption{\textbf{Evaluation on bimanual X-VLA across three RoboTwin~2.0 tasks} ($N{=}128$ for \textit{adjust\_bottle}, and $N{=}100$ for \textit{beat\_block\_hammer} and \textit{place\_empty\_cup}). Net texture reduction $G = P - A$ ($\mathrm{pp}$) and exact paired McNemar $p$-values ($p_{\text{Plain}}$) are measured against \textit{Plain}, separating the baseline carrier shift ($D = O - P = 32.03, 18.00,$ and $34.00\,\mathrm{pp}$). On \textit{adjust\_bottle} and \textit{beat\_block\_hammer}, \textit{RefCloth}, \textit{RefCloth-RT}, and \textit{X-VLA-Self} all significantly outperform \textit{Random~101} ($p \le .00904$, with $p \le 2.72{\times}10^{-8}$ for \textit{RefCloth} and $p \le 1.00{\times}10^{-11}$ for \textit{X-VLA-Self}), while on \textit{adjust\_bottle}, an additional control \textit{Random~102} attains $54.69\%$ ($p_{\text{Plain}}{=}.845$) and all three learned textures surpass it ($p \le 6.94{\times}10^{-8}$).}
\label{tab:rt_xvla}
\footnotesize
\setlength{\tabcolsep}{4.5pt}
\begin{tabular}{@{}lccccccccc@{}}
\toprule
& \multicolumn{3}{c}{\textit{adjust\_bottle} ($N{=}128$)} & \multicolumn{3}{c}{\textit{beat\_block\_hammer} ($N{=}100$)} & \multicolumn{3}{c}{\textit{place\_empty\_cup} ($N{=}100$)} \\
\cmidrule(lr){2-4}\cmidrule(lr){5-7}\cmidrule(lr){8-10}
Condition & Rate (\%) & $G\uparrow$ (pp) & $p_{\text{Plain}}$ & Rate (\%) & $G\uparrow$ (pp) & $p_{\text{Plain}}$ & Rate (\%) & $G\uparrow$ (pp) & $p_{\text{Plain}}$ \\
\midrule
Original ($O$)    & 85.16 & ---   & --- & 75.00 & ---   & --- & 53.00 & ---   & --- \\
Plain ($P$)       & 53.12 &  0.00 & --- & 57.00 &  0.00 & --- & 19.00 &  0.00 & --- \\
\midrule
Random~101        & 46.09 &  7.03 & .136 & 42.00 & 15.00 & .0275 &  6.00 & 13.00 & $9.77{\times}10^{-4}$ \\
\midrule
RefCloth (\cref{tab:main_transfer}) & \textbf{14.06} & \textbf{39.06} & $7.64{\times}10^{-11}$ &  \textbf{2.00} & \textbf{55.00} & $8.05{\times}10^{-16}$ & \textbf{3.00} & \textbf{16.00} & $4.02{\times}10^{-4}$ \\
RefCloth-RT       & 31.25 & 21.88 & $8.36{\times}10^{-6}$  & 26.00 & 31.00 & $3.35{\times}10^{-7}$  & \textbf{3.00} & \textbf{16.00} & $1.45{\times}10^{-4}$ \\
X-VLA-Self        &  \textbf{3.12} & \textbf{50.00} & $1.82{\times}10^{-18}$ &  \textbf{1.00} & \textbf{56.00} & $2.78{\times}10^{-17}$ & \textbf{3.00} & \textbf{16.00} & $8.55{\times}10^{-4}$ \\
\bottomrule
\end{tabular}
\end{table*}

\paragraph{Carrier efficacy versus single-source cross-backbone transfer on LIBERO.}
\Cref{tab:libero_xvla} isolates cross-backbone generalization on LIBERO without simulator change. Because X-VLA attains near-ceiling clean accuracy on LIBERO ($97.65\%$ \textit{Plain} mean across $2{,}000$ episodes), zero-shot transfer from a single OpenVLA-OFT source (DINOv2 and SigLIP) yields a moderate four-suite mean reduction of $G{=}6.25\,\mathrm{pp}$ ($91.40\%$ vs.\ $97.20\%$ for \textit{Random~101}), while still producing highly significant drops on Object ($G{=}9.80\,\mathrm{pp}$, $p{=}4.62{\times}10^{-14}$), Long ($G{=}11.80\,\mathrm{pp}$, $p{=}1.79{\times}10^{-9}$), and Goal ($G{=}3.00\,\mathrm{pp}$, $p{=}.0107$). Crucially, optimizing directly on X-VLA (\textit{X-VLA-Self}, $N{=}500$ per suite) collapses four-suite mean success to $20.75\%$ ($415/2{,}000$, $G{=}76.90\,\mathrm{pp}$ net drop, driving Spatial to $0.40\%$, Object to $12.40\%$, Goal to $18.40\%$, and Long to $51.80\%$, with $p \le 6.89{\times}10^{-53}$ vs.\ \textit{Plain} and $p \le 1.25{\times}10^{-36}$ vs.\ \textit{RefCloth} in every suite). Optimizing solely against X-VLA's Florence-2 VLM features (\textit{VLM-only}) achieves a nearly identical mean reduction ($19.25\%$, $G{=}78.40\,\mathrm{pp}$), confirming that the physical tablecloth carrier remains highly potent against a DaViT-based VLM backbone and that the smaller LIBERO zero-shot drop reflects single-source cross-backbone gradient alignment under high-certainty clean execution rather than carrier insensitivity.

\paragraph{Vulnerability of RL-hardened $\pi_{0.5}$ on RoboTwin~2.0.}
Turning to SAPIEN-based RoboTwin~2.0, \cref{tab:rt_ppo_bottle} evaluates whether RL post-training (\textbf{RLinf-$\pi_{0.5}$-PPO}) immunizes continuous policies against workspace surface perturbations. Relative to \textit{Plain} ($82.03\%$, reflecting a carrier shift of $D{=}10.16\,\mathrm{pp}$ from the $92.19\%$ \textit{Original} baseline), frozen LIBERO \textit{RefCloth} induces a statistically significant net reduction of $G{=}14.84\,\mathrm{pp}$ ($67.19\%$, $p{=}5.46{\times}10^{-4}$) and lowers success below both matched random controls (\textit{Random~101} at $71.88\%$ and \textit{Random~102} at $72.66\%$). Re-optimizing on single-task SAPIEN observations of the same policy family (\textit{RefCloth-RT}) deepens the disruption further to $65.62\%$ ($G{=}16.41\,\mathrm{pp}$, $p{=}4.92{\times}10^{-5}$), while RoboTwin-trained \textit{X-VLA-Self} drives success down to $56.25\%$ ($G{=}25.78\,\mathrm{pp}$, $p{=}3.61{\times}10^{-8}$ vs.\ \textit{Plain} and $p \le 8.21{\times}10^{-4}$ against both random controls), showing that RL fine-tuning does not eliminate surface-induced visual vulnerability.

\paragraph{Zero-shot cross-platform, cross-embodiment, and cross-backbone transfer on RoboTwin~2.0.}
Finally, \cref{tab:rt_xvla} evaluates zero-shot transfer across three bimanual RoboTwin~2.0 tasks on \textbf{X-VLA}. Because X-VLA is trained without tablecloth domain randomization, placing the unoptimized \textit{Plain} carrier induces substantial baseline carrier shifts $D$ ($32.03\,\mathrm{pp}$ on \textit{adjust\_bottle}, $18.00\,\mathrm{pp}$ on \textit{beat\_block\_hammer}, and $34.00\,\mathrm{pp}$ on \textit{place\_empty\_cup}), underscoring the necessity of our $D/G$ decomposition in \cref{eq:metrics}. Even after discounting this carrier shift, the single frozen \textit{RefCloth} PNG learned solely on single-arm OpenVLA-OFT in MuJoCo LIBERO transfers zero-shot with striking severity, collapsing success on \textit{adjust\_bottle} from $53.12\%$ (\textit{Plain}) and $46.09\%/54.69\%$ (\textit{Random~101/102}) to $14.06\%$ ($G{=}39.06\,\mathrm{pp}$, $p{=}7.64{\times}10^{-11}$ vs.\ \textit{Plain}, $p{=}2.72{\times}10^{-8}$ vs.\ \textit{Random~101}, and $p{=}3.07{\times}10^{-12}$ vs.\ \textit{Random~102}) and on \textit{beat\_block\_hammer} from $57.00\%$ (\textit{Plain}) and $42.00\%$ (\textit{Random~101}) to $2.00\%$ ($G{=}55.00\,\mathrm{pp}$, $p{=}8.05{\times}10^{-16}$ vs.\ \textit{Plain} and $p{=}1.96{\times}10^{-11}$ vs.\ \textit{Random~101}), closely approaching the domain-matched self-attack bound (\textit{X-VLA-Self} at $3.12\%$ and $1.00\%$). On \textit{place\_empty\_cup}, where \textit{Plain} already drops to $19.00\%$, all optimized patterns compress performance further to the $3.00\%$ floor ($G{=}16.00\,\mathrm{pp}$, $p \le 8.55{\times}10^{-4}$ vs.\ \textit{Plain}). Notably, whereas single-task \textit{RefCloth-RT} slightly edges out frozen \textit{RefCloth} within the $\pi_{0.5}$ family (\cref{tab:rt_ppo_bottle}), frozen multi-task LIBERO \textit{RefCloth} transfers substantially better than single-task \textit{RefCloth-RT} ($31.25\%$ and $26.00\%$) to the heterogeneous Florence-2 backbone of X-VLA, demonstrating that multi-task action-token displacement generalizes simultaneously across simulators, embodiments, and non-SigLIP VLM backbones without retraining.

\section{Dual-View Exposure and Behavior}
\label{supp:motivation}
Having established closed-loop failure rates across benchmarks, we next examine the physical and internal mechanisms behind these failures, tracing how the workspace cloth projects into native dual-view inputs, redirects source attention and action commands, and destabilizes continuous 3D end-effector trajectories.

\subsection{Background Support and Reference Offset}

\paragraph{Full-rollout background exposure.}
To quantify how consistently the cloth enters visual observations during closed-loop execution, we measure its native pixel footprint across 30 Plain rollouts per target--suite condition (\cref{tab:supp_exposure}). In tabletop setups (Spatial, Goal, Long), main-camera support exceeds $1\%$ at $100\%$ of execution steps, whereas in the floor-placed Object suite, wrist-view support ($34.36\%$--$35.72\%$) remains nearly three times larger than main-view coverage ($11.86\%$--$11.95\%$), confirming that the moving wrist camera maintains substantial surface exposure regardless of third-person vantage points.

\begin{table*}[t]
\centering\small\setlength{\tabcolsep}{5pt}
\caption{\textbf{Native Plain--hidden RGB support over complete rollouts.} Thirty Plain episodes per target--suite. Support reflects the median of episode medians (\%), with interquartile ranges in brackets. $q_{>1\%}$ is the median fraction of queries with main-view support $>1\%$.}
\label{tab:supp_exposure}
\begin{tabular}{@{}llrrrrr@{}}\toprule
Target & Suite & Successes / 30 & Queries / episode & Main support (\%) & Wrist (\%) & $q_{>1\%}$\\\midrule
$\pi_0$ & Spatial & 26 & 29.5 & 23.48 [22.86,24.48] & 41.27 & 1.000\\
 & Object & 26 & 36.9 & 11.86 [11.42,12.19] & 35.72 & 1.000\\
 & Goal & 22 & 39.4 & 22.41 [21.42,22.50] & 35.59 & 1.000\\
 & Long & 11 & 71.1 & 17.93 [15.03,23.57] & 33.38 & 1.000\\
\midrule
$\pi_{0.5}$ & Spatial & 30 & 22.2 & 23.72 [23.12,24.61] & 39.66 & 1.000\\
 & Object & 29 & 31.2 & 11.95 [11.55,12.15] & 34.36 & 1.000\\
 & Goal & 30 & 24.5 & 22.41 [21.55,22.49] & 37.88 & 1.000\\
 & Long & 23 & 59.1 & 18.70 [15.32,23.80] & 22.84 & 1.000\\
\bottomrule
\end{tabular}
\end{table*}

\paragraph{Unchanged-cloth reference offset.}
Alongside pixel coverage, we quantify the static discrepancy between differentiable and native rendering paths by evaluating the unperturbed base cloth ($T_0$) at 32 frozen Spatial states (\cref{fig:deployment}b). The relative feature offset $q(x)=\|f_R(0;x)-f_N(0;x)\|_2 / \|f_N(0;x)\|_2$ averages $16.02\%$ (spanning $8.58\%$ to $25.74\%$), showing that measuring displacement against native base renders introduces a large non-zero bias before optimization even begins.

\subsection{Dual-View Measurements at Actual Policy Inputs}
\label{supp:object_audit}
To confirm that these surface perturbations survive image resizing and normalization, we audit preprocessed $224\times224$ policy input tensors along 30 Object trajectories per target model (\cref{tab:supp_object_inputs}). Across $2{,}043$ policy queries, RefCloth alters $34.47\%$--$36.01\%$ of wrist-view input pixels with an RMS of $22.92$--$23.20$ (compared with $11.98\%$--$12.01\%$ and $10.69$--$10.70$ RMS in the main view), confirming that wrist observations inject the majority of adversarial perturbation into the vision encoders.

\begin{table*}[t]
\centering\small\setlength{\tabcolsep}{4.5pt}
\caption{\textbf{Same-state measurements at deployed Object policy inputs.} Summarizing 30 complete Plain trajectories per target at $224\times224$ resolution with threshold 10 on the 0--255 RGB scale. RMS measures median RefCloth--Plain pixel deviation.}
\label{tab:supp_object_inputs}
\begin{tabular}{@{}llrrrr@{}}\toprule
Target & View & Marker support & Plain--hidden & RefCloth--Plain & RMS\\\midrule
$\pi_0$ & Main & 14.58 [14.14,14.91] & 11.75 [11.27,12.05] & 11.98 [11.51,12.17] & 10.70\\
 & Wrist & 39.04 [32.50,52.38] & 36.00 [30.49,48.07] & 36.01 [31.02,47.02] & 23.20\\
\midrule
$\pi_{0.5}$ & Main & 14.63 [14.29,14.88] & 11.85 [11.41,12.03] & 12.01 [11.67,12.17] & 10.69\\
 & Wrist & 37.79 [31.99,45.06] & 34.67 [29.81,41.13] & 34.47 [29.77,39.39] & 22.92\\
\bottomrule
\end{tabular}
\end{table*}

\paragraph{Paired native closed-loop check.}
Evaluating Plain and RefCloth under strictly synchronized query-indexed sampling noise across 30 paired Object episodes (\cref{tab:supp_object_paired}) further removes rollout stochasticity. RefCloth induces 15 new failures on $\pi_0$ and 20 on $\pi_{0.5}$ with zero recovered episodes ($p_H\le6.10\times 10^{-5}$), verifying that closed-loop disruption is not driven by random action sampling.

\begin{table*}[t]
\centering\small\setlength{\tabcolsep}{6pt}
\caption{\textbf{Paired Object evaluation under synchronized sampling noise.} Thirty paired rollouts per target model. Induced/recovered counts measure Plain-success/RefCloth-failure transitions and the converse.}
\label{tab:supp_object_paired}
\begin{tabular}{@{}lrrrrrr@{}}\toprule
Target & Plain & RefCloth & $G$ (pp) & Induced & Recovered & $p_H$\\\midrule
$\pi_0$ & 26/30 & 11/30 & 50.0 & 15 & 0 & $6.10\!\times\!10^{-5}$\\
$\pi_{0.5}$ & 29/30 & 9/30 & 66.7 & 20 & 0 & $3.81\!\times\!10^{-6}$\\
\bottomrule
\end{tabular}
\end{table*}

\paragraph{Predeclared qualitative cases.}
Inspecting fixed Object evaluation rollouts reveals concrete target-selection errors on $\pi_{0.5}$, where RefCloth misleads the gripper into grasping incorrect distractor items (placing a green bottle instead of soup, and milk instead of ketchup), illustrating how learned surface patterns misdirect object-level manipulation in closed loop.

\paragraph{Target closed-loop camera-input interventions.}
To isolate how third-person and wrist cameras individually and jointly govern these closed-loop failures, we evaluate Plain, Main-only, Wrist-only, and Both across all 10 tasks and initial states 10--19 on $\pi_{0.5}$-Object ($100$ paired episodes per condition, $400$ total, \cref{tab:supp_camera_intervention} and \cref{fig:supp_mechanism}c). All four conditions share task--initial-state keys and synchronized query-indexed denoising noise, while the hybrid conditions (Main-only and Wrist-only) render both Plain and RefCloth at the exact camera state of each query and substitute only the designated view, with Plain and Both verified to produce bit-identical inputs and actions to uninstrumented rollouts.

As detailed in \cref{tab:supp_camera_intervention}, all six pairwise contrasts remain statistically significant after Holm correction ($p_H\le 0.00885$). Perturbing only the wrist view lowers success by $23$\,pp (discordant pairs $24/1$), more than doubling the $10$\,pp drop from perturbing only the main view ($11/1$, direct Main-only vs.\ Wrist-only contrast $+13$\,pp, discordant pairs $16/3$, $p_H=0.00885$). Moreover, exposing both views simultaneously (Both, $G=64$\,pp) exceeds the sum of single-view drops by $31$\,pp, as wrist perturbation reduces success by $23$\,pp when the main view is Plain but by $54$\,pp when the main view is RefCloth ($p_H=7.91\times 10^{-15}$). Among the $98$ Plain-successful episodes, $38$ succeed under both Main-only and Wrist-only individually yet fail under Both, confirming that dual-view projection eliminates single-view recovery fallbacks.

\begin{table}[t]
\centering\footnotesize
\setlength{\tabcolsep}{2.5pt}
\caption{\textbf{Pairwise contrasts for closed-loop interventions on $\pi_{0.5}$-Object.} 100 paired episodes per condition (Plain 98, Main 88, Wrist 75, Both 34). $\Delta$ denotes success difference (pp), Pairs counts discordant episodes (first / second succeeds), and $p_H$ is the Holm-adjusted exact McNemar $p$-value across all 6 contrasts.}
\label{tab:supp_camera_intervention}
\begin{tabular}{@{}lrrrr@{}}
\toprule
Pairwise contrast & $\Delta$ (pp) & Pairs & Exact $p$ & $p_H$\\
\midrule
Plain vs.\ Main & $+10$ & 11 / 1 & .00635 & .00885\\
Plain vs.\ Wrist & $+23$ & 24 / 1 & $1.6\times 10^{-6}$ & $4.7\times 10^{-6}$\\
Plain vs.\ Both & $+64$ & 64 / 0 & $1.1\times 10^{-19}$ & $6.5\times 10^{-19}$\\
Main vs.\ Wrist & $+13$ & 16 / 3 & .00443 & .00885\\
Main vs.\ Both & $+54$ & 55 / 1 & $1.6\times 10^{-15}$ & $7.9\times 10^{-15}$\\
Wrist vs.\ Both & $+41$ & 41 / 0 & $9.1\times 10^{-13}$ & $3.6\times 10^{-12}$\\
\bottomrule
\end{tabular}
\end{table}

\subsection{Fixed-State Source View Response}
\label{supp:source_views}
Complementing the closed-loop target interventions above, we next probe how learned textures register inside the frozen source policy across eight fixed states from Spatial tasks $\{0,2,6,8\}$, holding instructions, proprioception, and camera geometry strictly identical.

\begin{figure*}[t]
\centering
\includegraphics[width=.82\textwidth]{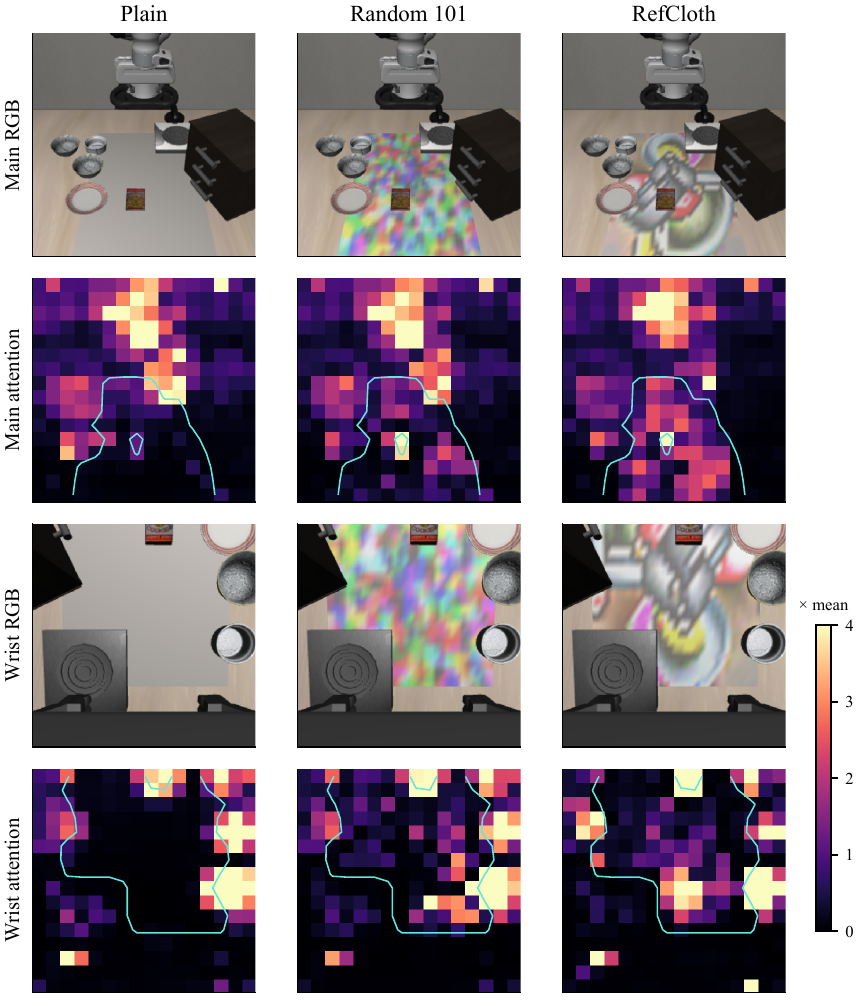}
\caption{\textbf{Source visual attention heatmaps at a representative state.} Spatial task 0, initial state 21, step 10. Native RGB inputs (rows 1 and 3) and corresponding $16\times16$ patch attention densities (rows 2 and 4). Cyan contours indicate visible cloth support.}
\label{fig:supp_source_attention}
\end{figure*}

\paragraph{Attention allocation.}
Measuring area-normalized action-token attention density $\rho_v$ (\cref{tab:supp_source_attention} and \cref{fig:supp_source_attention}) shows that both learned textures (Native ref. and RefCloth) concentrate substantially higher attention on the cloth carrier in the main view ($\rho_v=1.467$ and $1.271$) than Plain ($0.678$) or random controls ($0.775$--$0.785$), while also elevating wrist-view attention density ($0.692$ and $0.705$ vs.\ $0.376$ for Plain).

\begin{table}[t]
\centering\small\setlength{\tabcolsep}{7pt}
\caption{\textbf{Source attention allocation across eight frozen states.} Median area-normalized cloth attention density $\rho_v$ ($\rho_v=1$ corresponds to uniform spatial allocation within that view).}
\label{tab:supp_source_attention}
\begin{tabular}{@{}lrr@{}}\toprule
Texture & Main & Wrist\\\midrule
Plain & 0.678 & 0.376\\
Random 101 & 0.775 & 0.583\\
Random 102 & 0.785 & 0.693\\
Random 103 & 0.775 & 0.557\\
Native ref. & 1.467 & 0.692\\
UV off & 0.884 & 0.516\\
RefCloth & 1.271 & 0.705\\
\bottomrule\end{tabular}\end{table}

\begin{figure}[t]
\centering
\includegraphics[width=\columnwidth]{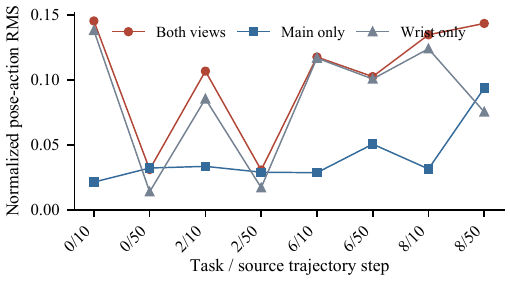}
\caption{\textbf{Two-view action attribution across eight frozen states.} Pose-action RMS drift relative to Plain under single-view or dual-view adversarial substitution. Connecting lines trace individual task steps.}
\label{fig:supp_source_actions}
\end{figure}

\paragraph{Two-view input interventions.}
Decomposing immediate pose-action disruption $\Delta a$ into symmetric main- and wrist-view contributions $(\phi_m,\phi_w)$ (\cref{fig:supp_source_actions}) further clarifies how each view acts across manipulation phases. At step 10, wrist-only substitution produces an action RMS of $0.116$ and accounts for an $85.0\%$ projection share of the total action shift, whereas by step 50 main and wrist views contribute comparably ($0.051$ vs.\ $0.052$ RMS), indicating that wrist observations dominate early table approach while third-person views govern terminal alignment.

\paragraph{Dual-view source gradient decomposition across training checkpoints.}
To verify that both camera projections actively shape the shared UV grid during optimization, we decompose the scaled feature-displacement gradient $\nabla_\theta\bigl(\lambda_f\mathcal D_{\mathrm{FD}}(\theta)\bigr) = g_m + g_w$ into main-camera ($g_m$) and wrist-camera ($g_w$) components by backpropagating each view's image gradient separately (verified against joint backpropagation with maximum relative residual $7.70\times 10^{-8}$) across checkpoints 0, 1,875, 3,750, and 5,000 and all four training seeds (32 batches per checkpoint, 128 batches total, \cref{tab:supp_source_grad_decomp}).

Across training (\cref{tab:supp_source_grad_decomp}), both cameras supply comparable gradient norms, rising from medians of $0.00089$ (main) and $0.00097$ (wrist) at step 0 to $0.06222$ and $0.06337$ at step 5,000. Notably, the two gradient vectors remain nearly orthogonal in parameter space (median cosine $0.005$--$0.054$), while the wrist projection share $s_w = \langle g_w, g_m+g_w\rangle / \|g_m+g_w\|_2^2$ stays balanced at $50.4\%$--$57.2\%$ across checkpoints (with step-5,000 seed medians of $59.1\%$, $40.4\%$, $51.1\%$, and $59.1\%$ on seeds 42, 7, 66, and 88), confirming that both views contribute complementary, non-conflicting updates from initialization to convergence.

\begin{table}[t]
\centering\footnotesize
\setlength{\tabcolsep}{3.2pt}
\caption{\textbf{Dual-view source gradient decomposition across training checkpoints.} Medians over 32 batches (4 seeds $\times$ 8 batches) of main-view ($\|g_m\|_2$) and wrist-view ($\|g_w\|_2$) control-grid gradient norms, inter-view cosine similarity $\cos(g_m, g_w)$, wrist projection share $s_w = \langle g_w, g_m+g_w\rangle/\|g_m+g_w\|_2^2$, and number of batches with $s_w > 50\%$.}
\label{tab:supp_source_grad_decomp}
\begin{tabular}{@{}rrrrrr@{}}\toprule
Step & $\|g_m\|_2$ & $\|g_w\|_2$ & $\cos(g_m,g_w)$ & $s_w$ (\%) & $s_w > 50\%$\\\midrule
0     & 0.00089 & 0.00097 & 0.005 & 53.7 & 19 / 32\\
1,875 & 0.04923 & 0.05757 & 0.054 & 57.2 & 23 / 32\\
3,750 & 0.06104 & 0.06564 & 0.030 & 52.7 & 18 / 32\\
5,000 & 0.06222 & 0.06337 & 0.046 & 50.4 & 16 / 32\\\bottomrule
\end{tabular}
\end{table}

\subsection{Continuous Trajectory Recordings}
\label{supp:continuous_trajectories}
\begin{figure*}[t]
\centering
\includegraphics[width=.98\textwidth]{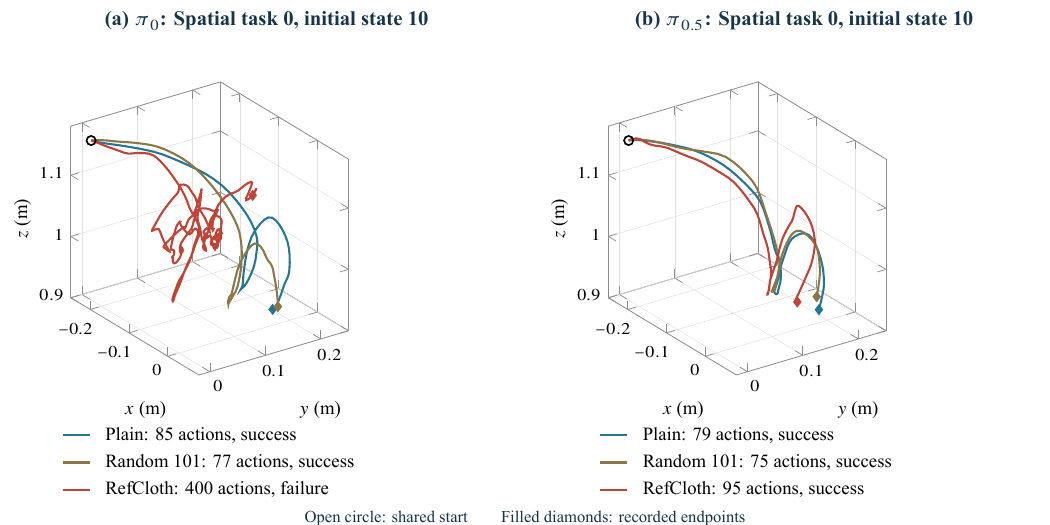}
\caption{\textbf{Three-dimensional continuous end-effector trajectories.} Continuous XYZ paths for Spatial task 0 under $\pi_0$ (a) and $\pi_{0.5}$ (b). Open circles mark shared starting positions, and filled diamonds indicate termination points.}
\label{fig:supp_continuous_paths}
\end{figure*}

Beyond binary task success, recording continuous 3D end-effector trajectories across six task--initial-state pairs (\cref{tab:supp_trajectories}) reveals severe dynamical degradation even in episodes that eventually succeed, such as $\pi_{0.5}$-Spatial task 2 requiring $348$ steps under RefCloth versus $92$ under Plain. As illustrated in \cref{fig:supp_continuous_paths}, RefCloth induces erratic vertical oscillations and detoured approach vectors that disrupt closed-loop control stability.

\begin{table*}[t]
\centering\small\setlength{\tabcolsep}{6pt}
\caption{\textbf{Continuous behavior recordings across representative tasks.} Executed action steps and outcomes (S denotes success, F denotes failure). Episodes share initial states across conditions.}
\label{tab:supp_trajectories}
\begin{tabular}{@{}lrrllllll@{}}\toprule
& & & \multicolumn{3}{c}{$\pi_0$} & \multicolumn{3}{c}{$\pi_{0.5}$}\\
\cmidrule(lr){4-6}\cmidrule(lr){7-9}
Suite & Task & Initial & Plain & Random & RefCloth & Plain & Random & RefCloth\\\midrule
Spatial & 0 & 10 & 85 S & 77 S & 400 F & 79 S & 75 S & 95 S\\
Spatial & 2 & 10 & 114 S & 110 S & 400 F & 92 S & 100 S & 348 S\\
Spatial & 0 & 12 & 75 S & 166 S & 400 F & 90 S & 82 S & 103 S\\
Goal & 5 & 10 & 125 S & 311 S & 400 F & 169 S & 126 S & 400 F\\
Goal & 7 & 10 & 81 S & 100 S & 400 F & 75 S & 78 S & 76 S\\
Goal & 5 & 12 & 144 S & 400 F & 400 F & 124 S & 114 S & 400 F\\\bottomrule
\end{tabular}
\end{table*}

\section{Theoretical Analysis}
\label{supp:theory}

Finally, we provide complete derivations and empirical verification for the three theoretical properties summarized in \cref{sec:matched_reference}, namely (i) global invariance of the matched reference to static rendering offsets and its residual error bound, (ii) the exact gradient discrepancy and its local impact on sign-gradient steps, and (iii) the quadratic local optimization geometry induced by surface transformations and multi-task sampling.

\subsection{Bias Invariance and Proxy Error Bound}
\label{supp:reference_proof}

Throughout this subsection, scene state $x$, camera views, instruction, proprioception, and UV transformation $u$ are held fixed, with $\theta=0$ corresponding to the unperturbed base texture $T_0$. Let $f_R(\theta)$ and $f_N(\theta)$ denote the flattened $m$-dimensional source action-token features under differentiable synthesis and native simulation, respectively.

We decompose the cross-renderer discrepancy into a static, texture-independent baseline offset $b$ and a texture-dependent residual $r_\theta$,
\begin{equation}
\label{eq:supp_error_decomp}
f_R(\theta) = f_N(\theta) + b + r_\theta, \qquad b = f_R(0) - f_N(0), \quad r_0 = 0.
\end{equation}
Relative to the true native texture-induced displacement $d^*(\theta) = f_N(\theta) - f_N(0)$, evaluating $f_R(\theta)$ against the native base reference $f_N(0)$ versus the path-matched reference $f_R(0)$ yields
\begin{equation}
\label{eq:supp_displacements}
\begin{aligned}
d_N(\theta) &= f_R(\theta) - f_N(0) = d^*(\theta) + b + r_\theta,\\
d_M(\theta) &= f_R(\theta) - f_R(0) = d^*(\theta) + r_\theta.
\end{aligned}
\end{equation}
Because \cref{eq:supp_displacements} holds identically for all $\theta$, same-path calibration cancels the static offset $b$ globally and recovers the exact native displacement ($d_M(\theta) = d^*(\theta)$) whenever renderer discrepancy is constant across textures ($r_\theta = 0$).

\paragraph{Bounded residual error.}
When texture-dependent rendering residuals are nonzero ($r_\theta \ne 0$), let $\mathcal D_{\mathrm{native}}^*(\theta) = \frac{1}{m}\|d^*(\theta)\|_2^2$ denote the ideal native displacement objective and $\mathcal D_{\mathrm{FD}}(\theta) = \frac{1}{m}\|d_M(\theta)\|_2^2$ our matched proxy objective. Whenever $\|d^*(\theta)\|_2 \le B$ and $\|r_\theta\|_2 \le \delta$ hold on the feasible parameter domain, applying the Cauchy--Schwarz and triangle inequalities gives
\begin{equation}
\label{eq:supp_proxy_bound}
\begin{aligned}
\bigl|\mathcal D_{\mathrm{FD}}(\theta) - \mathcal D_{\mathrm{native}}^*(\theta)\bigr|
&= \frac{1}{m}\bigl|2d^*(\theta)^\top r_\theta + \|r_\theta\|_2^2\bigr| \\
&\le \frac{2B\delta + \delta^2}{m}.
\end{aligned}
\end{equation}
This proxy error bound depends solely on the texture-induced residual $\delta$ and is completely independent of the static path offset $\|b\|_2$, which would otherwise inject $\frac{1}{m}(2(d^*(\theta)+r_\theta)^\top b + \|b\|_2^2)$ into the native-reference objective $\mathcal D_N(\theta)$.

\paragraph{Local estimation dominance near initialization.}
Near initialization ($\theta=0$), where $r_\theta$ is locally Lipschitz ($\|r_\theta\|_2 \le L_e\|\operatorname{vec}(\theta)\|_2$), the displacement estimation errors $E_M(\theta)=\|d_M(\theta)-d^*(\theta)\|_2$ and $E_N(\theta)=\|d_N(\theta)-d^*(\theta)\|_2$ satisfy
\begin{equation}
E_M(\theta) \le L_e\|\operatorname{vec}(\theta)\|_2, \qquad E_N(\theta) \ge \|b\|_2 - L_e\|\operatorname{vec}(\theta)\|_2.
\end{equation}
Hence $E_M(\theta) < E_N(\theta)$ holds strictly whenever $2L_e\|\operatorname{vec}(\theta)\|_2 < \|b\|_2$, explaining why path calibration is particularly decisive near the unperturbed base cloth where $\|b\|_2$ dominates early texture variations.

\begin{figure*}[t]
\centering
\includegraphics[width=.96\textwidth]{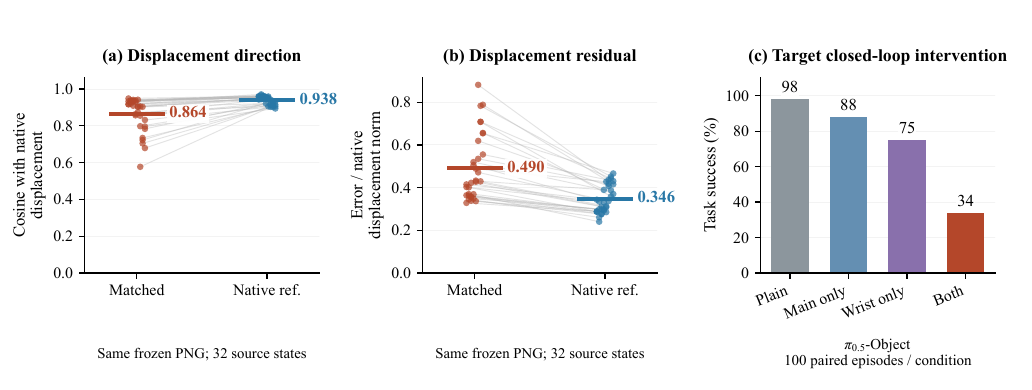}
\caption{\textbf{Trained-texture displacement fidelity and target closed-loop camera interventions.} (a) Cosine similarity and (b) relative displacement error ($\|d-d^*\|_2/\|d^*\|_2$) of matched ($d_M$) and native-reference ($d_N$) source feature displacements relative to true native displacement ($d^*$) across 32 Spatial states for the same frozen RefCloth PNG. (c) Closed-loop task success (\%) on $\pi_{0.5}$-Object across 100 paired episodes per camera-intervention condition.}
\label{fig:supp_mechanism}
\end{figure*}

\paragraph{Empirical measurement of trained displacement direction and residual.}
Algebraic cancellation of the static baseline offset $b$ is distinct from minimizing the texture-dependent residual $r_\theta$ at the converged texture $\theta^*$. To audit trained-texture fidelity directly, we compute $d_M(\theta^*)=f_R(T)-f_R(T_0)$, $d_N(\theta^*)=f_R(T)-f_N(T_0)$, and the true native displacement $d^*(\theta^*)=f_N(T)-f_N(T_0)$ for the same frozen RefCloth PNG across the 32 Spatial states from \cref{fig:deployment}b, both on the archived feature cache and on 32 freshly captured state-normalized snapshots verified for bit-exact repeatability (\cref{tab:supp_displacement_fidelity} and \cref{fig:supp_mechanism}a--b).

Across all 32 states, matched displacement $d_M$ preserves strong positive directional alignment with true native displacement $d^*$ (mean cosine $0.8644$ on fresh snapshots and $0.8612$ on archived states), with an average relative residual norm $\|r_\theta\|_2/\|d^*\|_2 = 0.4900$. For the same frozen texture, $d_N = d^* + b + r_\theta$ exhibits slightly higher cosine ($0.9382$) and lower relative error ($0.3459$) at convergence because static offset $b$ partially cancels $r_\theta$ on the high-contrast learned pattern, confirming that same-path calibration acts primarily by removing the zero-point bias $b$ that corrupts early gradient directions near $\theta=0$ (\cref{fig:full_gradients}).

\begin{table}[t]
\centering
\caption{\textbf{Displacement fidelity across 32 Spatial states.} Comparing matched ($d_M$) and native ($d_N$) displacements against true native displacement $d^*$ for frozen RefCloth. Rel.\ error is $\|d-d^*\|_2/\|d^*\|_2$ (which equals $\|r_\theta\|_2/\|d^*\|_2$ for Matched).}
\label{tab:supp_displacement_fidelity}
\resizebox{\columnwidth}{!}{%
\footnotesize
\setlength{\tabcolsep}{2.8pt}
\begin{tabular}{@{}lrrrr@{}}
\toprule
 & \multicolumn{2}{c}{Direction $\cos(d, d^*)$} & \multicolumn{2}{c}{Rel.\ Error $\|d-d^*\|/\|d^*\|$} \\
\cmidrule(lr){2-3}\cmidrule(l){4-5}
Snapshot (32 states) & Matched & Native ref. & Matched & Native ref. \\
\midrule
Archived (\cref{fig:deployment}b) & 0.8612 & 0.9393 & 0.5029 & 0.3437 \\
Fresh (PNG) & 0.8644 & 0.9382 & 0.4900 & 0.3459 \\
Fresh (float) & 0.8643 & 0.9381 & 0.4899 & 0.3460 \\
\bottomrule
\end{tabular}%
}
\end{table}

\subsection{Reference-Dependent Gradients and Local Updates}
\label{supp:sign_proof}

Setting $d_\theta = d_M(\theta) = z_\theta - z_0$, the cross-path squared norm expands identically for every $\theta$ as $\|d_\theta+b\|_2^2 = \|d_\theta\|_2^2 + 2d_\theta^\top b + \|b\|_2^2$. Differentiating with respect to $\operatorname{vec}(\theta)$ yields the exact gradient discrepancy
\begin{equation}
\label{eq:supp_gradient_difference}
\nabla_\theta\mathcal D_N - \nabla_\theta\mathcal D_{\mathrm{FD}} = \frac{2}{m}J_\theta^\top b,
\end{equation}
where $J_\theta=\partial z_\theta/\partial\operatorname{vec}(\theta)$, and taking expectations over $(x,u)$ under shared regularizers gives $g_N - g_M = \frac{2\lambda_f}{m}\mathbb E_{x,u}[J_\theta^\top b]$.

To see how this discrepancy alters sign-gradient ascent, consider $g_M=\nabla_\theta\mathcal J_M$ and $g_N=\nabla_\theta\mathcal J_N$ evaluated at a common parameter vector $\theta$ (such as our small initialization $\mathcal N(0, 0.01^2 I)$). When coordinates are nonzero and steps $\delta_M=\eta\operatorname{sign}(g_M)$ and $\delta_N=\eta\operatorname{sign}(g_N)$ are feasible, let $\mathcal S=\{i:\operatorname{sign}(g_{M,i})\ne\operatorname{sign}(g_{N,i})\}$ denote the sign-disagreement set. The linearized gain advantage on $\mathcal J_M$ is then
\begin{equation}
\label{eq:supp_sign_advantage}
\begin{aligned}
g_M^\top(\delta_M-\delta_N)
&= \eta\sum_i g_{M,i}\bigl[\operatorname{sign}(g_{M,i})-\operatorname{sign}(g_{N,i})\bigr]\\
&= 2\eta\sum_{i\in\mathcal S}|g_{M,i}| = 2\eta\|g_M\|_1\rho_{\mathrm{flip}},
\end{aligned}
\end{equation}
where $\rho_{\mathrm{flip}}=\sum_{i\in\mathcal S}|g_{M,i}|/\|g_M\|_1$. Under box bounds $\ell\le\theta_i\le r$, the coordinate-wise constrained linear maximizer over $\|\delta\|_\infty\le\eta$ takes the form $\delta_i^*=\operatorname{clip}(\eta\operatorname{sign}(g_{M,i}),\ell-\theta_i,r-\theta_i)$, and whenever $\mathcal J_M$ has an $L$-Lipschitz gradient in $d$ dimensions, Taylor's theorem gives
\begin{equation}
\label{eq:supp_finite_step}
\mathcal J_M(\theta+\delta_M)-\mathcal J_M(\theta+\delta_N) \ge 2\eta\sum_{i\in\mathcal S}|g_{M,i}| - L\eta^2d.
\end{equation}

\paragraph{Measured one-step gain.}
Empirically, at step size $\eta=0.05$, the matched update improves $\mathcal J_M$ by $0.0024695$ more than the native update on the initial mini-batch, closely matching the first-order prediction of $0.0023480$ from \cref{eq:supp_sign_advantage} ($5.2\%$ relative difference). Across 20 paired batches, gradient cosine similarity between $g_M$ and $g_N$ rises from $0.010$ at initialization to $0.980$ at convergence, confirming that $J_\theta^\top b$ primarily distorts early optimization steps near the base cloth.

\subsection{Local Optimization Geometry and Multi-Pose Sampling}
\label{supp:sampling_proof}

Finally, we characterize the local curvature induced by shared UV augmentation around $\theta=0$. Letting $A_{x,u} = \left.\partial z_\theta(x,u)/\partial\operatorname{vec}(\theta)\right|_{\theta=0}$ and using $d_0(x,u)=0$, the first-order expansion $d_\theta(x,u) \approx A_{x,u}\operatorname{vec}(\theta)$ yields the local quadratic form
\begin{equation}
\mathcal L_P(\theta) = \mathbb E_{x,u\sim P}[\mathcal D_{\mathrm{FD}}(\theta;x,u)] \approx \operatorname{vec}(\theta)^\top H_P \operatorname{vec}(\theta),
\end{equation}
where the expected sensitivity Gram matrix is
\begin{equation}
H_P = \frac{1}{m}\mathbb E_{x,u\sim P}\bigl[A_{x,u}^\top A_{x,u}\bigr],
\end{equation}
whereas the uncalibrated native objective retains the linear bias $\frac{2}{m}\mathbb E_{x,u}[b(x)^\top A_{x,u}]\operatorname{vec}(\theta) + \frac{1}{m}\mathbb E_x[\|b(x)\|_2^2]$ at $\theta=0$.

Because each state $x$ satisfies the variance decomposition $\mathbb E_u\|d_\theta(x,u)\|_2^2 = \|\mathbb E_u d_\theta(x,u)\|_2^2 + \mathbb E_u\|d_\theta(x,u)-\mathbb E_u d_\theta(x,u)\|_2^2$, maximizing $\mathcal L_P(\theta)$ rewards texture directions with large average squared response across surface poses $u\sim P$ (governed locally by $H_P$) without requiring feature displacements to remain strictly invariant across poses. For a mixture of source tasks with weights $p_k$ and gradients $\nabla_\theta\mathcal J_k(\theta)$, the common first-order sign update maximizing $\sum_k p_k\nabla_\theta\mathcal J_k^\top\delta$ under $\|\delta\|_\infty\le\eta$ is $\delta=\eta\operatorname{sign}(\bar g)$ with $\bar g=\sum_k p_k\nabla_\theta\mathcal J_k(\theta)$, achieving linearized gain $\eta\|\bar g\|_1$.
\end{document}